\documentclass[10pt,twocolumn,letterpaper]{article}

\usepackage{cvpr}              
\makeatletter
\@namedef{ver@everyshi.sty}{}
\makeatother

\usepackage{cvpr}
\usepackage{booktabs}
\usepackage{times}
\usepackage{epsfig}
\usepackage{graphicx}
\usepackage{amsmath}
\usepackage{amssymb}
\usepackage{array}
\usepackage{multirow}
\usepackage[font=small,labelfont=bf]{caption}
\usepackage{cite}
\usepackage{enumitem}            
\usepackage{array, makecell} %
\usepackage{caption}
\usepackage{adjustbox}
\usepackage[table]{xcolor}       
\definecolor{mygray}{gray}{0.95}
\usepackage{hhline}
\usepackage{nicematrix}
\usepackage{xr-hyper}
\usepackage[pagebackref,breaklinks,colorlinks]{hyperref}
\usepackage[capitalize]{cleveref}
\crefname{section}{Sec.}{Secs.}
\Crefname{section}{Section}{Sections}
\Crefname{table}{Table}{Tables}
\crefname{table}{Tab.}{Tabs.}

\newcommand{\vect}[1]{\boldsymbol{#1}}
\newcommand{\minisection}[1]{\vspace{1mm} \noindent \textbf{#1} \hspace{0mm}}
\newcommand{\xiyin}[1]{{\color{blue}[\textbf{Xi:} #1]}}
\newcommand{\va}[1]{{\color{red}[\textbf{Vishal:} #1]}}
\newcommand{\xl}[1]{{\color{green}[\textbf{Xiaoming:} #1]}}

\usepackage{xcolor,pifont}
\newcommand*\colourcheck[1]{%
  \expandafter\newcommand\csname #1check\endcsname{\textcolor{#1}{\ding{52}}}%
}

\newcommand*\colourcross[1]{%
  \expandafter\newcommand\csname #1check\endcsname{\textcolor{#1}{\ding{54}}}%
}

\colourcheck{green}
\colourcross{red}

\setlist[itemize]{noitemsep,topsep=0pt,leftmargin=*,label={\large\textbullet}}

\newcommand{\Section}[1]{\vspace{0.5mm} \section{#1} \vspace{0.5mm}}
\newcommand{\SubSection}[1]{\vspace{0.5mm} \subsection{#1} \vspace{0.5mm}}
\newcommand{\SubSubSection}[1]{\vspace{0.5mm} \subsubsection{#1} \vspace{0.5mm}}

\usepackage[capitalize]{cleveref}
\crefname{section}{Sec.}{Secs.}
\Crefname{section}{Section}{Sections}
\Crefname{table}{Table}{Tables}
\crefname{table}{Tab.}{Tabs.}

\def\cvprPaperID{4766} 
\def\confName{CVPR}
\def\confYear{2023}

\begin{document}

\title{MaLP: Manipulation Localization Using a Proactive Scheme}

\author{
Vishal Asnani$^{1}$, \,\,
Xi Yin$^{2}$, \,\,
Tal Hassner$^{2}$, \,\,
Xiaoming Liu$^{1}$ \\
$^1$\thanks{All data sourcing, modeling codes, and experiments were developed at Michigan State University. Meta did not obtain the data/codes or conduct any experiments in this work. } Michigan State University,\,\,$^2$Meta AI\\
{\tt\small$^1$\{asnanivi, liuxm\}@msu.edu,\,\,$^2$\{yinxi, thassner\}@meta.com}}
\maketitle

\begin{abstract}
   Advancements in the generation quality of various Generative Models (GMs) has made it necessary to not only perform binary manipulation detection but also localize the modified pixels in an image. However, prior works termed as \textit{passive} for manipulation localization exhibit poor generalization performance over unseen GMs and attribute modifications. To combat this issue, we propose a \textit{proactive} scheme for manipulation localization, termed MaLP. We encrypt the real images by adding a learned template. If the image is manipulated by any GM, this added protection from the template not only aids binary detection but also helps in identifying the pixels modified by the GM. The template is learned by leveraging local and global-level features estimated by a two-branch architecture. We show that MaLP performs better than prior passive works. We also show the generalizability of MaLP by testing on $22$ different GMs, providing a benchmark for future research on manipulation localization. 
   Finally, we show that MaLP can be used as a discriminator for improving the generation quality of GMs. Our models/codes are available at \url{www.github.com/vishal3477/pro\_loc}.
\end{abstract}

\section{Introduction}

We witness numerous Generative Models (GMs)~\cite{goodfellow2014generative, tran2017disentangled,liu2019stgan, choi2018stargan, park2019gaugan, CycleGAN2017, stargan2,most-gan-3d-morphable-stylegan-for-disentangled-face-image-manipulation, karras2019style,karras2018progressive, wang2021sketch, gu2022vector, rombach2022high, kim2022diffusionclip} being proposed to generate realistic-looking images.
These GMs can not only generate an entirely new image~\cite{karras2019style,karras2018progressive}, but also perform partial manipulation of an input image~\cite{stargan2, liu2019stgan, stargan2, CycleGAN2017}. 
The proliferation of these GMs has made it easier to manipulate personal media for malicious use.
Prior methods to combat manipulated media focus on binary detection
~\cite{ pscc-net-progressive-spatio-channel-correlation-network-for-image-manipulation-detection-and-localization,shiohara2022detecting, gerstner2022detecting, chen2022self, xu2022supervised, asnani2022proactive, rossler2019faceforensics, wu2020sstnet, dang2020detection,asnani2021reverse}, using mouth movement, model parsing, hand-crafted features,~\etc.

Recent works go one step further than detection, \ie \textit{manipulation localization}, which is defined as follows: given a partially manipulated image by a GM (\eg STGAN~\cite{liu2019stgan} modifying hair colors of a face image), the goal is to identify which pixels are modified by estimating a \textit{fakeness map}~\cite{huang2022fakelocator}. 
Identifying modified pixels helps to determine the severity of the fakeness in the image, and aid  media-forensics~\cite{huang2022fakelocator, dang2020detection}. Also, manipulation localization provides an understanding of the attacker's intent for modification which may further benefit identifying attack toolchains used~\cite{darpa}. 

Recent methods for manipulation localization~\cite{li2020face, songsri2019complement,nguyen2019multi} focus on estimating the manipulation mask of face-swapped images. They localize modified facial attributes by leveraging attention mechanisms~\cite{dang2020detection}, patch-based classifier~\cite{chai2020makes}, and face-parsing~\cite{huang2022fakelocator}. 
The main drawback of these methods is that they do not generalize well to GMs unseen in training.
That is when the test images and training images are modified by different GMs, 
which  will likely happen given the vast number of existing GMs.
Thus, our work aims for a localization method generalizable to unseen GMs. 

\begin{figure}[t!]
\centering
\includegraphics[trim={0 -4 0 0},clip,width=\columnwidth]{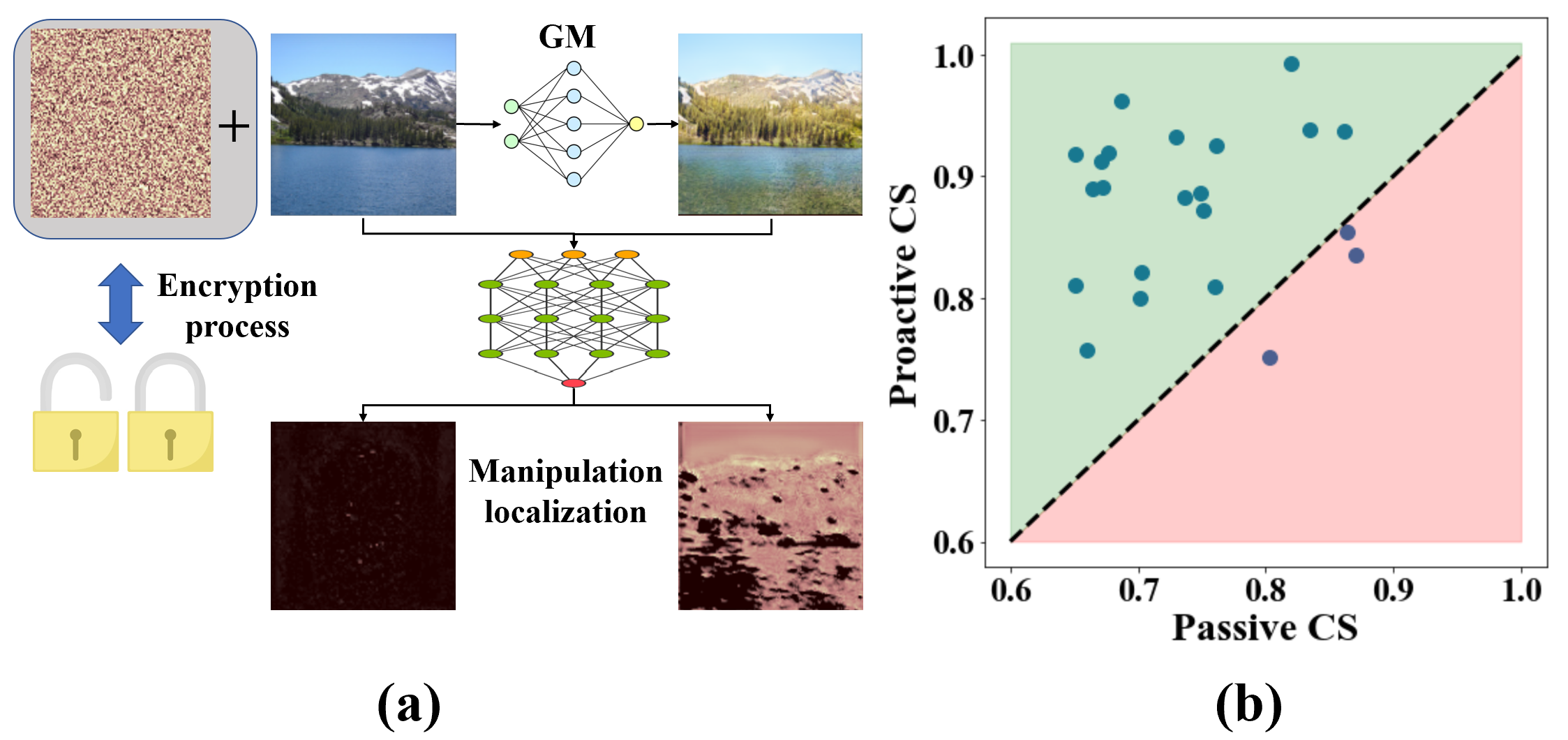}
\caption{\textbf{(a)} \textbf{High-level idea of MaLP.} We encrypt the image by adding a learnable template, which helps to estimate the fakeness map. \textbf{(b)} The cosine similarity (CS) between ground-truth and predicted fakeness maps for $22$ unseen GMs. The performance is better for almost all GMs when using our proactive approach.}
\label{fig:teaser}
\vspace{-3mm}
\end{figure}

All aforementioned methods are based on a \textit{passive} scheme as the method receives an image as is for estimation.
Recently, proactive methods are gaining success for deepfake tasks such as detection~\cite{asnani2022proactive}, disruption~\cite{ruiz2020disrupting, yeh2020disrupting}, and tagging~\cite{wang2021faketagger}. 
These methods are considered \textit{proactive} as they add different types of  signals known as \textit{templates} for encrypting the image before it is manipulated by a GM. This template can be one-hot encoding~\cite{wang2021faketagger}, adversarial perturbation~\cite{ruiz2020disrupting}, or a learnable noise~\cite{asnani2022proactive}, and is optimized to improve the performance of the defined tasks. 

Motivated by~\cite{asnani2022proactive}, we propose a Proactive scheme for MAnipulation Localization, termed as MaLP, in order to improve generalization. 
Specifically, MaLP learns an optimized template which, when added to real images, would improve manipulation localization, should they get manipulated. This manipulation can be done by an unseen GM trained on either in-domain or out-of-domain datasets. 
Furthermore, face manipulation may involve modifying facial attributes unseen in training (\eg train on hair color modification yet test on  gender modification). 
MaLP incorporates three modules that focus on encryption, detection, and localization. 
The encryption module selects and adds the template from the template set to the real images. 
These encrypted images are further processed by localization and detection modules to perform the respective tasks. 
Designing a proactive manipulation localization approach comes with several challenges. First, it is not straightforward to formulate constraints for learning the template {\it unsupervisedly}. 
Second, calculating a fakeness map at the same resolution as the input image is computationally expensive if the decision for each pixel has to be made. 
Prior works~\cite{chai2020makes, dang2020detection} either down-sample the images or use a patch-wise approach, both of which result in inaccurate low-resolution fakeness maps. 
Lastly, the templates should be generalizable to localize modified regions from unseen GMs. 

We design a two-branch architecture consisting of a shallow CNN network and a transformer to optimize the template during training. 
While the former leverages local-level features due to its shallow depth, the latter focuses on global-level features to better capture the affinity of the far-apart regions. 
The joint training of both networks enables the MaLP to learn a better template, having embedded the information of both levels. 
During inference, the CNN network alone is sufficient to estimate the fakeness map with a higher inference efficiency.
Compared to prior passive works~\cite{huang2022fakelocator, dang2020detection}, MaLP improves the generalization performance on unseen GMs. 
We also demonstrate that MaLP can be used as a discriminator for fine-tuning conventional GMs to improve the quality of GM-generated images.

In summary, we make the following contributions.
\begin{itemize}
    \item We are the first to propose a proactive scheme for image manipulation localization, applicable to both face and generic images.
    \item Our novel two-branch architecture uses both local and global level features to learn a set of templates in an unsupervised manner. The framework is guided by constraints based on template recovery, fakeness maps classification, and high cosine similarity between predicted and ground-truth fakeness maps. 
    \item MaLP can be used as a plug-and-play discriminator module to fine-tune the generative model to improve the quality of the generated images.
    \item Our method outperforms State-of-The-Art (SoTA) methods in manipulation localization and detection. Furthermore, our method generalizes well to GMs and modified attributes unseen in training. To facilitate the research of localization, we develop a benchmark for evaluating the generalization of manipulation localization, on images where the train and test GMs are different.
\end{itemize}

\begin{table}[t]
\centering
\caption{Comparison of our approach with prior works on manipulation localization and proactive schemes. We show the generalization ability of all works across different facial attribute modifications, unseen GMs trained on datasets with the same domain (in-domain) and different domains (out-domain). [Keys:  Attr.: Attributes, Imp.: Improving, L.: Localization, D.: Detection]
}
\begin{adjustbox}{width=1\columnwidth}
\begin{NiceTabular}{c|c|c|c|c|c|c|c}
\hline\hline
\rowcolor{mygray}  &  &  &  & \multicolumn{3}{c|}{Generalization}   & Imp.\\ \hhline{~|~|~|~|-|-|-|~}
\rowcolor{mygray} \multirow{-2}{*}{Work}& \multirow{-2}{*}{Scheme}& \multirow{-2}{*}{Task}& \multirow{-2}{*}{Template}& Attr. & In-domain & Out-domain & GM\\\hline
\cite{wang2021faketagger} & Proactive & Tag & Fix & \greencheck & \greencheck & \redcheck & \redcheck\\
\rowcolor{mygray}\cite{segalis2020ogan} & Proactive & Disrupt & Learn & \greencheck & \redcheck & \redcheck & \redcheck\\
\cite{ruiz2020disrupting} & Proactive & Disrupt & Learn & \greencheck & \greencheck & \redcheck & \redcheck\\
\rowcolor{mygray}\cite{yeh2020disrupting} & Proactive & Disrupt & Learn & \greencheck & \redcheck & \redcheck & \redcheck\\
\cite{asnani2022proactive} & Proactive & D. & Learn & \redcheck & \greencheck & \greencheck & \redcheck\\
\rowcolor{mygray}\cite{nguyen2019multi} & Passive & L. $+$ D. & -& \redcheck & \redcheck & \redcheck & \redcheck\\
\cite{songsri2019complement} & Passive & L. $+$ D. &-&  \redcheck & \redcheck & \redcheck & \redcheck\\
\rowcolor{mygray}\cite{li2020face} & Passive & L. $+$ D. & -& \redcheck & \greencheck & \redcheck & \redcheck\\
\cite{dang2020detection} & Passive & L. $+$ D. & -& \greencheck & \greencheck & \redcheck & \redcheck\\
\rowcolor{mygray}\cite{chai2020makes} & Passive & L. $+$ D. & -& \redcheck & \greencheck & \redcheck & \redcheck\\
\cite{huang2022fakelocator} & Passive & L. $+$ D. &-&  \greencheck & \greencheck & \redcheck & \redcheck\\\hline
\rowcolor{mygray}MaLP & Proactive & L. $+$ D. & Learn & \greencheck& \greencheck& \greencheck& \greencheck\\
 \hline \hline
\end{NiceTabular}
\label{tab:rel_works}
\end{adjustbox}
\vspace{-3mm}
\end{table}

\section{Related Work}
\minisection{Manipulation Localization.}
Prior works tackle manipulation localization by adopting a passive scheme. Some of them focus on forgery attacks like removal, copy-move, and splicing using multi-task learning~\cite{nguyen2019multi}. Songsri-in~{\it et~al.}~\cite{songsri2019complement} leverage facial landmarks~\cite{chollet2017xception} for manipulation localization. Li~\etal~\cite{li2020face} estimate the blended boundary for forged face-swap images. ~\cite{dang2020detection} uses an attention mechanism to leverage the relationship between pixels and \cite{chai2020makes} uses a patch-based classifier to estimate modified regions. 
Recently, Huang~\etal~\cite{huang2022fakelocator} utilize gray-scale maps as ground truth for manipulation localization and leverage face parsing with an attention mechanism for prediction. 
The passive methods discussed above suffer from the generalization issue~\cite{nguyen2019multi, chollet2017xception, songsri2019complement, dang2020detection, chai2020makes, huang2022fakelocator} and estimate a low-resolution fakeness map~\cite{dang2020detection} which is less accurate for the localization purpose. 
MaLP generalizes better to modified attributes and GMs unseen in training.

\minisection{Proactive Scheme.}
Recently, proactive schemes are developed for various tasks.
Wang~\etal~\cite{wang2021faketagger} leverage the recovery of embedded one-hot encoding messages to perform deepfake tagging. A small perturbation is added onto the images by Segalis~\etal~\cite{segalis2020ogan} to disrupt the output of a GM. The same task is performed by Ruiz~\etal~\cite{ruiz2020disrupting} and Yeh~\etal~\cite{yeh2020disrupting}, both adding adversarial noise onto the input images. Asnani~\etal~\cite{asnani2022proactive} propose a framework based on adding a learnable template to input images for generalized manipulation detection. Unlike prior works, which focus on binary detection, deepfake disruption, or tagging, our work emphasizes on manipulation localization. We show the comparison of our approach with prior works in Tab.~\ref{tab:rel_works}. 


\minisection{Manipulation Detection.}
The advancement in manipulation detection keeps reaching new heights. Prior works propose to combat deepfakes by exploiting frequency domain patterns~\cite{wang2020cnn}, up-sampling artifacts~\cite{zhang2019detecting}, model parsing~\cite{asnani2021reverse, yao2022reverse}, hand-crafted features~\cite{nataraj2019detecting}, lip motions~\cite{rossler2019faceforensics}, unified detector~\cite{deb2023unified} and self-attention~\cite{dang2020detection}. Recent methods use self-blended images~\cite{shiohara2022detecting}, hierarchical localization features~\cite{hierarchical-fine-grained-image-forgery-detection-and-localization}, real-time deviations~\cite{gerstner2022detecting}, and self-supervised learning with adversarial training~\cite{chen2022self}. Finally, methods based on contrastive learning~\cite{xu2022supervised} and proactive scheme~\cite{asnani2022proactive} have explicitly focused on generalized manipulation detection across unknown GMs.  

\Section{Proposed Approach}

\begin{figure*}[t!]
\centering
\includegraphics[width=0.98\textwidth]{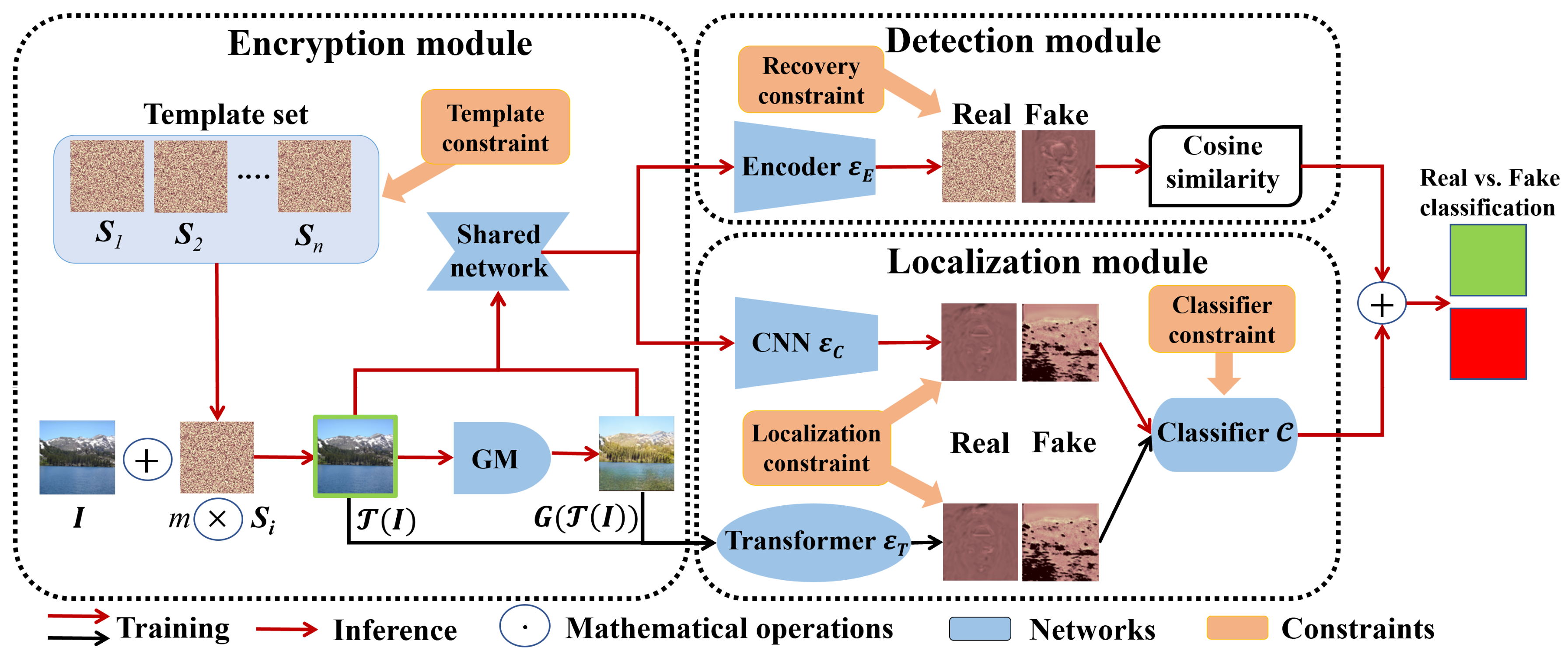}
\vspace{0.5mm}
\caption{ \textbf{The overview of MaLP.} It includes three modules: encryption, localization, and detection. We randomly select a template from the template set and add it to the real image as encryption. The GM is used in inference mode to manipulate the encrypted image. The detection module recovers the added template for binary detection. The localization module uses a two-branch architecture to estimate the fakeness map. Lastly, we apply the classifier to the fakeness map to better distinguish them from each other. Best viewed in color.}
\label{fig:overview}
\vspace{-3mm}
\end{figure*}

\SubSection{Problem Formulation}
\minisection{Passive Manipulation Localization}
Let $\vect{I}^R$ be a set of real images that are manipulated by a GM $G$ to output the set of manipulated images $G(\vect{I}^R)$. 
Prior passive works perform manipulation localization by estimating the fakeness  map $\vect{M}_{pred}$ with the following objective:
{\small
\begin{equation}
    \min_{\theta_{\mathcal{E}}}\bigg\{\sum_j \Big (\Big |\Big |\mathcal{E}(G(\vect{I}^R_j);\theta_{\mathcal{E}})-\vect{M}_{GT} \Big |\Big |_2\Big ) \bigg\},
    \label{eq:pred_map_passive}
\end{equation}
}
where $\mathcal{E}$ denotes the passive framework with parameters $\theta_{\mathcal{E}}$ and $\vect{M}_{GT}$ is the ground-truth fakeness map.

To represent the fakeness map, some prior methods~\cite{songsri2019complement, dang2020detection, li2020face} choose a binary map by applying a threshold on the difference between the real and manipulated images. This is undesirable as the threshold selection is highly subjective and sensitive, leading to inaccurate fakeness maps. 
Therefore, we adopt the continuous gray-scale map for calculating the ground-truth fakeness maps~\cite{huang2022fakelocator}, formulated as:
{\small
\begin{equation}
    \vect{M}_{GT}= Gray(|\vect{I}^R-G(\vect{I}^R)|)/255,
    \label{eq:gt_passive}
\end{equation}
}
where $Gray(.)$ converts the image to gray-scale. 

\minisection{Proactive Scheme}
Asnani~\etal~\cite{asnani2022proactive} define adding the template as a transformation $\mathcal{T}$ applied to images $\vect{I}^R$, resulting in the  encrypted images $\mathcal{T}(\vect{I}^R)$. The added template acts as a signature of the defender and is learned during the training, aiming to improve the performance of the task at hand, \eg detection, disruption, and tagging. 
Motivated by~\cite{asnani2022proactive} that uses multiple templates, we have a set of $n$ orthogonal templates ${\bf\mathcal{S}}=\{\vect{S}_1, \vect{S}_2,...\vect{S}_n\}$ where $\vect{S}_i$ $\in \mathbb{R}^{128\times128}$, for a real image $\vect{I}_j^R\in \vect{I}^R$, transformation $\mathcal{T}$ is defined as:
{\small
\begin{equation}
    \mathcal{T}(\vect{I}_j^R;\vect{S}_i)={\vect{I}_j^R} + {\vect{S}_i},  \text{ where }i\in \{1,2,...,n\}.
    \label{eq:add_temp}
\end{equation}
}
The templates are optimized such that adding them to the real images wouldn't result in a noticeable visual difference, yet helps manipulation localization.

\minisection{Proactive Manipulation Localization.} 
Unlike the passive schemes~\cite{huang2022fakelocator, dang2020detection, nguyen2019multi, li2020face}, we learn an optimal template set to help manipulation localization. For the encrypted images $\mathcal{T}(\vect{I}^R)$, we formulate the estimation of the fakeness map as:
{\small
\begin{equation}
    \min_{\theta_{\mathcal{E}_P},\vect{S}_i}\bigg\{\sum_j \Big (\Big |\Big|\mathcal{E}_P(G(\mathcal{T}(\vect{I}^R_j;\vect{S}_i));\theta_{\mathcal{E}_P})-\vect{M}_{GT} \Big |\Big |_2\Big ) \bigg\}.
    \label{eq:pred_map_proactive}
\end{equation}
}
where $\mathcal{E}_P$ is the proactive framework with parameters $\theta_{\mathcal{E}_P}$.

However, as the output of the GM has changed from images in set $G(\vect{I}^R)$ to images in set $G(\mathcal{T}(\vect{I}^R))$, in our proactive approach, the calculation of the ground-truth fakeness map shall be changed from Eq.~\ref{eq:gt_passive} to the follows:
{\small
\begin{equation}
    \vect{M}_{GT}= Gray(|\vect{I}^R-G(\mathcal{T}(\vect{I}^R))|)/255.
    \label{eq:gt_proactive}
\end{equation}
}

\SubSection{Manipulation Localization}
MaLP consists of three modules: encryption, localization, and detection. The encryption module is used to encrypt the real images. The localization module estimates the fakeness map using a two-branch architecture. The detection module performs binary detection for the encrypted and manipulated images by recovering the template and using the classifier in the localization module. All three modules, as detailed next, are trained in an end-to-end manner. 

\SubSubSection{Encryption Module}
Following the procedure in~\cite{asnani2022proactive}, we add a randomly selected learnable template from the template set to a real image. We control the strength of the added template using a hyperparameter $m$, which  prevents the degradation of the image quality. The encryption process is summarised below:
{\small
\begin{equation}
    \mathcal{T}({\vect{I}^R_j})={\vect{I}^R_j} + m \times {\vect{S}_i} \text{ where}\hspace{1mm} i= Rand(1,2,...,n).
\end{equation}
}We select the value of $m$ as $30\%$ for our framework. 

\begin{figure}[t!]
\centering
\includegraphics[width=\columnwidth]{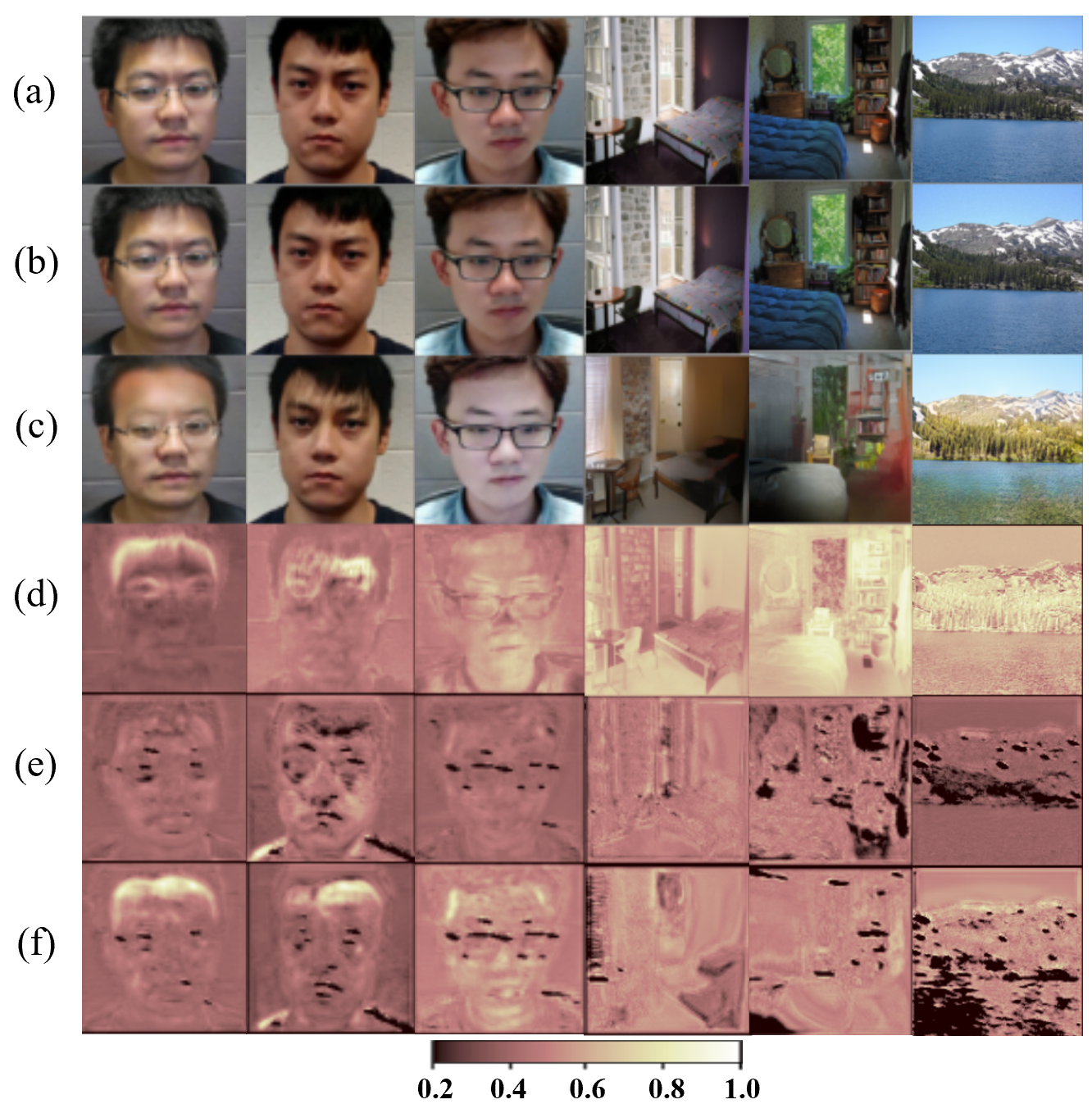}
\caption{Visualization of fakeness maps for faces and generic images showing generalization across unseen attribute modifications and GMs: (a) real image, (b) encrypted image, (c) manipulated image, (d) $\vect{M}_{GT}$, (e) predicted fakeness map for encrypted images, and (f) predicted fakeness map for manipulated images. 
The first column shows the manipulation of (seen GM, seen attribute modification)~\ie (STGAN, bald). Following two columns show the manipulation of (seen GM, unseen attribute modification)~\ie (STGAN, [bangs, pale skin]. The fourth and fifth columns show manipulation of unseen GM, GauGAN for non-face images. The last column shows manipulation by unseen GM, DRIT. We see that the fakeness map of manipulated images is more bright and similar to  $\vect{M}_{GT}$, while the real fakeness map is more close to zero. We use the cmap as ``pink" to better visualize the fakeness map. All face images come from SiWM-v2 data~\cite{swim}. }
\label{fig:vis}
\vspace{-3mm}
\end{figure}

We optimize the template set by focusing on properties like low magnitude, orthogonality, and high-frequency content~\cite{asnani2022proactive}. The properties are applied as constraints as follows. 
{\small
\begin{equation}
J_T = \lambda_1\times\sum_{i=1}^n||{\vect{S}_i}||_2 + \lambda_2\times\sum_{\substack{i,j=1\\ i \neq j}}^n\text{CS}(\vect{S}_i,\vect{S}_j) +  \lambda_3\times||\mathcal{L}(\mathfrak{F}({\vect{S}}))||_2,
\label{eq:Jt}
\end{equation}
}
where CS is the cosine similarity, $\mathcal{L}$ is the low-pass filter, $\mathfrak{F}$ is the fourier transform, $\lambda_1$, $\lambda_2$, $\lambda_3$ are weights for losses of low magnitude, orthogonality and high-frequency content, respectively.

\SubSubSection{Localization Module}
\label{sec:localization}
To design the localization module, we consider two desired properties: a larger receptive field for fakeness map estimation and high inference efficiency. 
A network with a large receptive field will consider far-apart regions in learning features for localization. 
Yet, large receptive fields normally come from deeper networks, implying slower inference. 

In light of these properties, we design a two-branch architecture consisting of a shallow CNN network $\mathcal{E}_C$ and a ViT transformer~\cite{dosovitskiy2020image} $\mathcal{E}_T$ (see Fig.~\ref{fig:overview}). 
The intuition is to have one shallow branch to capture local features, and one deeper branch to capture global features.
While training with both branches helps to learn better templates, in inference we only use the shallow branch for a higher efficiency.
Specifically, the shallow CNN network has $10$ layers which is efficient in inference but can only capture the local features due to small receptive fields.
To capture global information, we adopt the ViT transformer. With the self-attention between the image patches, the transformer can estimate the fakeness map considering the far-apart regions. 

Both the CNN and transformer are trained jointly to estimate a better template set, resembling the concept of the ensemble of networks. 
We  empirically show that training both networks simultaneously results in higher performance than training either network separately. 
As the shallow CNN network is much faster in inference than the transformer,  we use the transformer only in training to optimize the templates and switch off the transformer branch in inference. 

To estimate the fakeness map, we leverage the supervision of the ground-truth fakeness map in Eq.~\ref{eq:gt_proactive}. For fake images, we maximize the cosine similarity ($CS$) and structural similarity index measure ($SS$) between the predicted and ground-truth fakeness map. However, the fakeness map should be a zero image for encrypted images. Therefore, we apply an $L_2$ loss~\cite{huang2022fakelocator} to minimize the predicted map to zero for encrypted images. 
To maximize the difference between the two fakeness maps, we further minimize the cosine similarity between the predicted map from encrypted images and $\vect{M}_{GT}$. 
The localization loss is defined as:
{\small
\begin{equation}
\label{eq:Jl}
J_L=
    \begin{cases}
        \Big\{\lambda_4\times||\mathcal{E}_{C/T}(\vect{I})||_2^2+ & \text{if } \vect{I}\in \mathcal{T}(\vect{I}^R)\\
        \lambda_5\times\text{CS}(\mathcal{E}_{C/T}(\vect{I}), \vect{M}_{GT})\Big\}\\
        \Big\{\lambda_6\times(1-\text{CS}(\mathcal{E}_{C/T}(\vect{I}), \vect{M}_{GT}))+ & \text{if } \vect{I} \in G(\mathcal{T}(\vect{I}^R))\\
        \lambda_7\times (1-SS(\mathcal{E}_{C/T}(\vect{I}), \vect{M}_{GT}))\Big\}.
    \end{cases}
\end{equation}
}
Finally, we have a classifier  to make a binary decision of real \vs fake using the fakeness maps. This classifier is included in the framework to aid the detection module for binary detection of the input images, which will be discussed in Sec.~\ref{sec:detection}. Another reason to have the classifier is to make the fakeness maps from encrypted and fake images to be distinguishable. We find that this design allows our training to converge much faster.

\SubSubSection{Detection Module}
\label{sec:detection}
To leverage the added template for manipulation detection, we perform template recovery using encoder $\mathcal{E}_E$. 
We follow the procedure in~\cite{asnani2022proactive} to recover the added template from the encrypted images by maximizing the cosine similarity between $\vect{S}$ and $\vect{S}_R$. However, for manipulated images, we minimize the cosine similarity between the recovered template ($\vect{S}_R$) and all the templates in the template set $\mathcal{S}$.
{\small
\begin{equation}
J_R=
    \begin{cases}
        \lambda_8\times(1-\text{CS}(\vect{S},\vect{S}_R)) & \text{if } x \in \mathcal{T}(\vect{I}^R)\\
        \lambda_9\times(\sum_{i=1}^n(\text{CS}(\vect{S}_i,\vect{S}_R))) & \text{if } x \in G(\mathcal{T}(\vect{I}^R)).
    \end{cases}
\end{equation}
}
Further, we leverage our estimated fakeness map to help manipulation detection. As discussed in the previous section, we apply a classifier $\mathcal{C}$ to perform binary classification of the predicted fakeness map for the encrypted and fake images. The logits of the classifier are further combined with the cosine similarity of the recovered template. The averaged logits are back-propagated using the binary cross-entropy constraint. This not only improves the performance of manipulation detection but also helps manipulation localization. 
Therefore, we apply the binary cross entropy loss on the averaged logits as follows:
{\small
\begin{equation}
\begin{aligned}
   J_C=&\lambda_{10}\times-\sum_j \bigg \{y_j. \text{log}\Big [\frac{\mathcal{C}({\vect{X}_j})+CS(\vect{S}_R, \vect{S})}{2}\Big]-\\
    &(1-y_j).\text{log}\Big[1-\frac{\mathcal{C}({\vect{X}_j})+CS(\vect{S}_R, \vect{S})}{2}\Big]\bigg \},
    \label{eq:binary_obj}
\end{aligned}
\end{equation}
}where $y_j$ is the class label, $\vect{S}$ and $\vect{S}_R$ are the added and recovered template respectively. 

Our framework is trained in an end-to-end manner with the overall loss function as follows:
{\small
\begin{equation}
J = {J_T}+{J_R}+{J_C}+{J_{L}}.
\label{eqn:multi_temp}
\end{equation}
}

\subsection{MaLP as A Discriminator}

One application of MaLP is to leverage our proposed localization module as a discriminator for improving the quality of the manipulated images. MaLP performs binary classification by estimating a fakeness map, which can be used as an objective. This results in output images being resilient to manipulation localization, thereby lowering the performance of our framework. 


We use MaLP as a plug-and-play discriminator to improve image generation quality through fine-tuning pretrained GMs. 
The generation quality and manipulation localization will compete head-to-head, resulting in a better quality of the manipulated images. 
We define the fine-tuning objective for the GM as follows:
{\small
\begin{align}
\begin{split}
    &\min_{\theta_G}\max_{\theta_{MaLP},\vect{S}_i}\bigg\{\sum_j \Big (\mathbb E\big[log(\mathcal{E}_{MaLP}(\mathcal{T}(\vect{I}^R_j));\theta_{MaLP})\big]+\\
    &\mathbb E\big[1-log(\mathcal{E}_{MaLP}(G(\mathcal{T}(\vect{I}^R_j;\vect{S}_i);\theta_{G});\theta_{MaLP}))\big] \Big ) \bigg\}.
\end{split}
    \label{eq:obj_disc}
\end{align}
}where $\mathcal{E}_{MaLP}$ is our framework with $\theta_{MaLP}$ parameters.

\section{Experiments}

\begin{table}[t]
\rowcolors{1}{mygray}{white}
\begin{center}
\small
\centering{
\caption{Manipulation localization comparison with prior works.}
\begin{adjustbox}{width=1\columnwidth}
\label{tab:baseline_comp_loc_det}
\begin{NiceTabular}{c|c|c|c|c|c|c}
\hline\hline
\rowcolor{mygray} & \multicolumn{3}{c|}{Localization}& \multicolumn{3}{c}{Detection}\\\hhline{~|-|-|-|-|-|-}
\rowcolor{mygray} \multirow{-2}{*}{Method}& CS $\bf \uparrow$ & PSNR $\bf \uparrow$ & SSIM $\bf \uparrow$& Accuracy $\bf \uparrow$& EER $\bf \downarrow$& AUC $\bf \uparrow$\\\hline
\cite{dang2020detection} & $0.6230$ & $6.214$ & $0.2178$ & $0.9975$ & $\bf0.0050$ & $0.9975$\\
\rowcolor{mygray}\cite{huang2022fakelocator} & $0.8831$ & $22.890$ & $\bf0.7876$ & $0.9945$ & $0.0077$ & $0.9998$\\
MaLP  & $\bf0.9394$ & $\bf 23.020$ & $0.7312$ & $\bf0.9991$ & $0.0072$ & $\bf1.0$\\\hline\hline
\end{NiceTabular}
\end{adjustbox}}
\end{center}
\vspace{-3mm}
\end{table}

\subsection{Experimental Setup}
\minisection{Settings} Following the settings in~\cite{huang2022fakelocator}, we use STGAN~\cite{liu2019stgan} to manipulate images from CelebA~\cite{liu2015faceattributes} dataset and train on bald facial attribute modification. 
In order to evaluate the generalization of image manipulation localization, we construct a new benchmark that consists of $200$ real images of $22$ different GMs on various data domains. 
The real images are chosen from the dataset on which the GM is trained on. 
The list of GMs, datasets and implementation details are provided in the supplementary.     




\minisection{Evaluation Metrics} 
We use cosine similarity (CS), peak signal-to-noise ratio (PSNR), and structural similarity index measure (SSIM) as adopted by~\cite{huang2022fakelocator} to evaluate manipulation localization since the GT is a continuous map. For binary detection, we use the area under the curve (AUC), equal error rate (EER), and accuracy score~\cite{huang2022fakelocator}. 


\begin{table}[t]
\rowcolors{1}{mygray}{white}
\centering
\small
\caption{Comparison of localization performance across unseen GMs and attribute modifications. We train on STGAN bald/smile attribute modification and test on AttGAN/StyleGAN.}
\begin{adjustbox}{width=1\columnwidth}
\label{tab:diff_attr}
\begin{NiceTabular}{c|c|c|c|c|c|c}
\hline\hline
\rowcolor{mygray} & \multicolumn{3}{c|}{Cosine similarity $\bf \uparrow$(AttGAN)}& \multicolumn{3}{c}{ Cosine similarity $\bf \uparrow$ (StyleGAN)}\\\hhline{~|-|-|-|-|-|-}
\rowcolor{mygray} \multirow{-2}{*}{Method} & Bald & Black Hair & Eyeglasses & Smile & Age & Gender\\\hline
\cite{huang2022fakelocator} & $0.8141$ & $0.6932$ & $0.6950$ & $0.6176$ & $0.3141$ & $0.6470$\\
\rowcolor{mygray}MaLP  & $\bf 0.8201$ & $\bf 0.7940$ & $\bf 0.8557$ & $\bf 0.8159$ & $\bf 0.8255$ & $\bf 0.8016$\\\hline\hline
\end{NiceTabular}
\end{adjustbox}
\vspace{-3mm}
\end{table}

\subsection{Comparison with Baselines}
We compare our results with~\cite{huang2022fakelocator} and~\cite{dang2020detection} for manipulation localization. The results are shown in Tab.~\ref{tab:baseline_comp_loc_det}. MaLP has higher cosine similarity and similar PSNR for localization compared to~\cite{huang2022fakelocator}. However, we observe a dip in SSIM. This might be because of the degradation caused by adding our template to the real images and then performing the manipulation. The learned template helps localize the manipulated regions better, as demonstrated by cosine similarity, but the degradation affects SSIM and PSNR. We also compare the performance of real \vs fake binary detection. As expected, our proposed proactive approach outperforms the passive methods with a perfect AUC and near-perfect accuracy. We also show visual examples of fakeness maps for images modified by unseen GMs in Fig.~\ref{fig:vis}. MaLP is able to estimate the fakeness map for unseen modifications and GMs across face/generic image datasets. 

\begin{table*}[t!]
\small
\centering
\caption{ Benchmark for manipulation localization across $22$ different unseen GMs, showing cosine similarity between ground-truth and predicted fakeness maps. We compare our proactive \vs passive baselines~\cite{chai2020makes,he2016deep, dang2020detection} approach to highlight the generalization ability of our MaLP. We scale the images to $128^2$ for ``sc." and keep the resolution as is for ``no sc.".}
\vspace{1.5mm}
\begin{adjustbox}{width=1\textwidth}
\begin{NiceTabular}{c|c|c|c|c|c|c|c|c|c|c|c|c}
\hline\hline
\rowcolor{mygray} GM & SEAN~\cite{Zhu_2020_sean}& StarGAN~\cite{choi2018stargan} & CycleGAN~\cite{CycleGAN2017} & GauGAN~\cite{park2019gaugan} & Con\_Enc.~\cite{pathakCVPR16context}  & StarGAN2~\cite{stargan2} & ALAE~\cite{pidhorskyi2020alae} & BiGAN~\cite{zhu2017bicycle} & AuGAN~\cite{CycleGAN2017} & GANim~\cite{pumarola2018ganimation} & DRGAN~\cite{tran2017disentangled} & ILVR~\cite{choi2021ilvr}\\\hline\hline
Resolution & $256^2$ & $128^2$ & $256^2$ & $256^2$ & $128^2$ & $256^2$ & $256^2$ & $256^2$ & $340^2$ & $128^2$ & $128^2$ & $256^2$ \\\hline
\rowcolor{mygray} ResNet50~\cite{he2016deep} & $0.8614$ & $0.7513$ & $0.6715$  & $0.7615$ & $\bf 0.8639$ & $0.8196$ & $0.6766$ & $0.6514$ & $0.6639$ & $0.6871$ &  $\bf 0.8029$ & $0.7018$\\
\cite{chai2020makes} & $0.7514$ & $0.7111$ & $0.7981$ & $0.8016$ & $0.7894$ & $0.7026$ & $0.7156$ & $0.7217$ & $0.7516$ & $0.7612$ & $0.7115$ & $0.7851$ \\
\rowcolor{mygray} \cite{dang2020detection} & $0.7961$ & $0.7887$ & $0.8014$ & $0.8256$ & $0.8541$ & $0.7034$ & $0.7549$ & $0.7805$ & $0.7232$ & $0.8457$ & $0.7239$ & $0.7854$ \\\hline
MaLP (sc.) & $\bf 0.9376$ & $\bf 0.8718$ & $0.9128$ & $\bf 0.9251$ & $0.8546$ & $\bf 0.8836$ & $\bf 0.9192$ & $0.9181$ & $ 0.8894$ & $\bf 0.9625$ & $0.7512$ & $0.8003$\\
\rowcolor{mygray} MaLP (no sc.)& $0.9258$ & $\bf0.8718$ & $\bf0.9245$ & $0.9125$ & $0.8546$ & $0.8785$ & $0.9141$ & $\bf 0.9229$ & $\bf 0.9149$ & $\bf 0.9625$ & $\bf0.7512$ & $\bf0.8359$     \\\hline\hline
GM & DRIT~\cite{DRIT} & Pix2Pix~\cite{isola2017pix2pix} & CounGAN~\cite{nizan2020council} & DualGAN~\cite{yi2018dualgan}  & ESRGAN~\cite{wang2021realesrgan} & UNIT~\cite{liu2018unit} & MUNIT~\cite{huang2018munit} & ColGAN~\cite{nazeri2018image} & GDWCT~\cite{GDWCT2019} & RePaint~\cite{lugmayr2022repaint} & \multicolumn{2}{c}{Average}\\\hline\hline
\rowcolor{mygray} Resolution & $256^2$ & $256^2$ & $128^2$ & $256^2$ & $1024^2$ & $512 \times 931$ & $256 \times 512$ & $128^2$ & $128^2$ & $256^2$ & \multicolumn{2}{c}{-}\\\hline
ResNet50~\cite{he2016deep} & $0.7486$  & $0.6719$  & $0.7293$ & $0.7365$ & $\bf 0.8703$ & $0.7083$ & $0.6601$ & $0.7596$ & $0.8350$ & $0.6512$ & \multicolumn{2}{c}{$0.7401$}\\
\rowcolor{mygray} \cite{chai2020makes} & $0.7871$ & $0.7769$ & $0.8146$ & $0.7569$ & $0.8168$ & $0.8064$ & $0.6788$ & $0.7610$ & $0.8691$ & $0.7516$ & \multicolumn{2}{c}{$0.7645$}\\ 
\cite{dang2020detection} & $0.8120$ & $0.7781$ & $0.8559$ & $0.7721$ & $0.8241$ & $0.8086$ & $0.7097$ & $0.7874$ & $0.8879$ & $0.7696$ & \multicolumn{2}{c}{$0.7903$} \\\hline
\rowcolor{mygray} MaLP (sc.)& $0.8867$ & $\bf 0.8915$ & $\bf 0.9326$ & $\bf 0.8872$ & $0.8348$ & $0.8214$ & $0.7565$ & $\bf 0.8096$ & $\bf 0.9384$ & $0.8102$ &\multicolumn{2}{c}{$0.8725$} \\
MaLP (no sc.) & $\bf 0.9084$ & $0.8714$ & $\bf0.9326$ & $0.8432$ & $\bf0.8743$ & $0.8391$ & $\bf0.7860$ & $\bf0.8096$ & $\bf0.9384$ & $\bf0.8290$ & \multicolumn{2}{c}{$\bf0.8773$}\\\hline\hline
\end{NiceTabular}
\label{tab:gm_gen}
\end{adjustbox}
\vspace{-3mm}
\end{table*}

\subsection{Generalization}
\label{sec:gen}
\minisection{Across Attribute Modifications}
Following the settings in~\cite{huang2022fakelocator}, we evaluate the performance of MaLP across unseen attribute modifications. Specifically, we train MaLP using STGAN with the bald/smile attribute modification and test it on unseen attribute modifications with unseen GMs: AttGAN/StyleGAN. 
As shown in Tab.~\ref{tab:diff_attr}, MaLP is more generalizable to all unseen attribute modifications. Furthermore, AttGAN shares the high-level architecture with STGAN but not with StyleGAN. We observe a significant increase in localization performance for StyleGAN compared to AttGAN. This shows that, unlike our MaLP, passive works perform much worse if the test GM doesn't share any similarity with the training GM.

\minisection{Across GMs}
Although~\cite{huang2022fakelocator} tries to show generalization across unseen GMs; it is limited by the GMs within the same domain of the dataset used in training. We propose a benchmark to evaluate the generalization performance for future manipulation localization works that consists of $22$ different GMs in various domains. We select GMs that are publicly released and can perform partial manipulation. 

As no open-source code base is available for~\cite{huang2022fakelocator}, we train a passive approach using a ResNet50~\cite{he2016deep} network to estimate the fakeness map as the baseline for comparison. Further, we compare our approach with~\cite{chai2020makes, dang2020detection}. Although~\cite{chai2020makes, dang2020detection} estimate a fakeness map, it has at least $5\times$ lower resolution compared to input images due to their patch-based methodology. 
For a fair comparison, we rescale their predicted fakeness maps to the resolution of $\vect{M}_{GT}$. We compare the cosine similarity in Tab.~\ref{tab:gm_gen}. MaLP is able to outperform all the baselines for almost all GMs, which proves the effectiveness of the proactive scheme.

We also evaluate the performance of $\mathcal{E}_C$ for high-resolution images. For encryption, we upsample the $128\times 128$ template to the original resolution of images and evaluate $\mathcal{E}_C$ on these higher resolution encrypted images. We observe similar performance of $\mathcal{E}_C$ for higher resolution images in Tab.~\ref{tab:gm_gen}, proving the versatility of $\mathcal{E}_C$ to image sizes. 

\subsection{Improving Quality of GMs}
\begin{figure}[t!]
\centering
\includegraphics[width=\columnwidth]{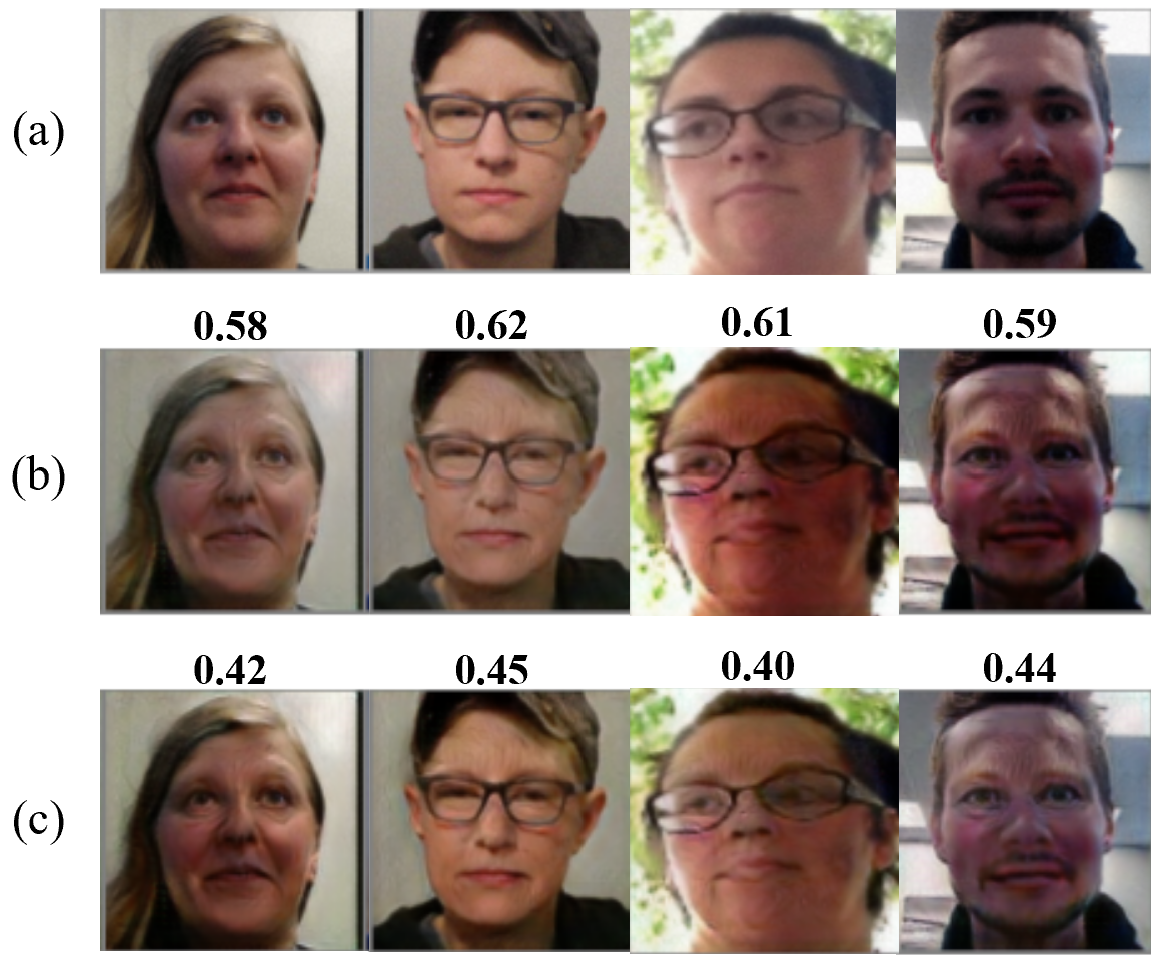}
\caption{ Visualization of (a) encrypted images, (b) manipulated images before fine-tuning, and (c) manipulated images after fine-tuning. The generation quality has improved after we fine-tune the GM using our framework as a discriminator. The artifacts in the images have been reduced, and the face skin color is less pale and more realistic. We also specify the cosine similarity of the predicted fakeness map and $\vect{M}_{GT}$. The GM is able to decrease the performance of our framework after fine-tuning. All face images come from SiWM-v2 data~\cite{swim}. }
\label{fig:gm_finetune}
\vspace{-3mm}
\end{figure}

We fine-tune the GM into fooling our framework to generate a fakeness map as a zero image. This process results in better-quality images. Initially, we train MaLP with the pretrained GM so that it can perform manipulation localization. Next, to fine-tune the GM, we adopt two strategies. First, we freeze MaLP and fine-tune the GM only. Second, we fine-tune both the GM and the MaLP but update the MaLP with a lower learning rate. The result for fine-tuning StarGAN is shown in Tab.~\ref{tab:disc_app}. We observe that for both strategies, MaLP reduces the FID score of StarGAN. We also show some visual examples in Fig.~\ref{fig:gm_finetune}. We see that the images are of better quality after fine-tuning, and many artifacts in the images manipulated by the pretrained model are removed.

\begin{table}[t]
\rowcolors{1}{mygray}{white}
\centering
\small
\caption{ FID score comparison for the application of our approach as a discriminator for improving the generation quality of the GM} 
\begin{adjustbox}{width=0.55\columnwidth}
\begin{NiceTabular}{c|c|c}
\hline\hline
\rowcolor{mygray}State&Fine-tune & StarGAN FID $\bf \downarrow$ \\\hline
Before & $-$ & $60.49$ \\\hline
\rowcolor{mygray} & $G$ & $\bf51.91$ \\
\rowcolor{mygray}\multirow{-2}{*}{After}& $G+MaLP$ & $52.07$\\\hline\hline
\end{NiceTabular}
\label{tab:disc_app}
\end{adjustbox}
\vspace{-3mm}
\end{table}

\subsection{Other Comparisons}
\minisection{Binary Detection}
We compare with prior proactive and passive approaches for binary manipulation detection~\cite{asnani2022proactive, wang2020cnn, nataraj2019detecting,zhang2019detecting}. We adopt the evaluation protocol in~\cite{asnani2022proactive} to test on images manipulated by CycleGAN, StarGAN, and GauGAN. We are able to perform similar to~\cite{asnani2022proactive} as shown in Tab.~\ref{tab:baseline_comp_proac}. We have better average precision than passive schemes and generalize well to GMs unseen in training.
We also conduct experiments to see whether localization can help binary detection to improve the performance, as mentioned in Sec.~\ref{sec:detection}. 
The combined predictions' results are better than just using the detection module as shown in Tab.~\ref{tab:baseline_comp_proac}. 
This is intuitive as the localization module provides extra information, thereby increasing the performance.

\minisection{Inference Speed} We compare the inference speed of our MaLP against prior work. ~\cite{huang2022fakelocator} uses Deeplabv3-ResNet101 model from PyTorch~\cite{paszke2019pytorch}. In our generalization benchmark shown in Sec.~\ref{sec:gen}, we use the ResNet50 model for training the passive baseline. The inference speed per image on an NVIDIA K$80$ GPU for Deeplabv3-ResNet101, ResNet50, and MaLP are $75.61$, $52.66$, and $29.26$ ms, respectively. MaLP takes less than half the inference time compared to~\cite{huang2022fakelocator} due to our shallow CNN network.

\begin{table}[t]
\rowcolors{1}{mygray}{white}
\small
\centering
\caption{ Comparison with prior binary detection works. [Keys: D.M.: Detection module, L.M.: Localization module] } 
\begin{adjustbox}{width=1\columnwidth}
\label{tab:baseline_comp_proac}
\begin{NiceTabular}{c|c|c|c|c|c}
\hline\hline
\rowcolor{mygray} &  & Set & \multicolumn{3}{c}{Test GM Average precision (\%)$\bf \uparrow$}\\\hhline{~|~|~|-|-|-}
\rowcolor{mygray}\multirow{-2}{*}{Method} & \multirow{-2}{*}{Train GM}&size& CycleGAN & StarGAN & GauGAN\\\hline
Nataraj~\etal~\cite{nataraj2019detecting} & CycleGAN & - & $\bf100$ & $88.20$ & $56.20$\\
\rowcolor{mygray}Zhang~\etal~\cite{zhang2019detecting} & AutoGAN &-& $\bf100$ & $\bf100$ & $61.00$\\
Wang~\etal~\cite{wang2020cnn} & ProGAN &-& $84.00$ & $\bf100$ & $67.00$\\
\rowcolor{mygray}Asnani~\etal~\cite{asnani2022proactive} & STGAN &$1$& $94.00$ & $\bf 100$ & $69.50$\\\hline
MaLP (D.M.) & STGAN & $1$ & $94.10$ & $\bf 100$ & $69.61$\\
\rowcolor{mygray}MaLP (D.M. $+$ L.M.) & STGAN & $1$ & $94.30$ & $\bf 100$ & $\bf72.16$ \\\hline\hline
\end{NiceTabular}
\end{adjustbox}
\vspace{-3mm}
\end{table}

\begin{table}[t]
\rowcolors{1}{mygray}{white}
\centering
\small
\caption{Comparison with adversarial attack methods.}
\begin{adjustbox}{width=1\columnwidth}
\label{tab:adv_comp}
\begin{NiceTabular}{c|c|c|c|c}
\hline\hline
 \rowcolor{mygray}&  & \multicolumn{3}{c}{Cosine similarity$\bf \uparrow$}\\\hhline{~|~|-|-|-}
\rowcolor{mygray}\multirow{-2}{*}{Method}& \multirow{-2}{*}{Scheme} & Bald & Black Hair & Eyeglasses\\\hline
Huang~\etal~\cite{huang2022fakelocator} &Passive& $0.8141$ & $0.6932$ & $0.6950$\\
\rowcolor{mygray}PGD~\cite{madry2018towards} & Proactive & $0.8051$ & $0.7514$ & $0.8358$\\
FGSM~\cite{goodfellow2014explaining} & Proactive & $0.8111$ & $0.7882$ & $0.8512$\\
\rowcolor{mygray}CW~\cite{carlini2017towards} & Proactive & $0.8014$ & $\bf 0.8344$ & $0.8405$\\
MaLP & Proactive & $\bf 0.8201$ & $0.7940$ & $\bf 0.8557$\\\hline\hline
\end{NiceTabular}
\end{adjustbox}
\vspace{-3mm}
\end{table}

\minisection{Adversarial Attack} Our framework can be considered as an adversarial attack on real images to aid manipulation localization. Therefore, it is vital to contrast the performance between our approach and classic adversarial attacks. For this purpose, we perform experiments that make use of adversarial attacks, namely PGD~\cite{madry2018towards}, CW~\cite{carlini2017towards}, and FGSM~\cite{goodfellow2014explaining} to guide the learning of the added template. We evaluate on unseen GM AttGAN for unseen attribute modifications. We show the performance comparison in Tab.~\ref{tab:adv_comp}. MaLP has higher cosine similarity across some unseen facial attribute modifications compared to adversarial attacks. This can be explained as the adversarial attack methods being over-fitted to training parameters (data, target network~\etc). Therefore, if the testing data is changed with unseen attribute modifications by GMs, the performance of adversarial attacks degrades. Further, these attacks are analogous to our MaLP as a proactive scheme which, in general, have better performance than passive works.

\begin{figure}[t!]
\rowcolors{1}{mygray}{white}
\centering
\includegraphics[width=\columnwidth]{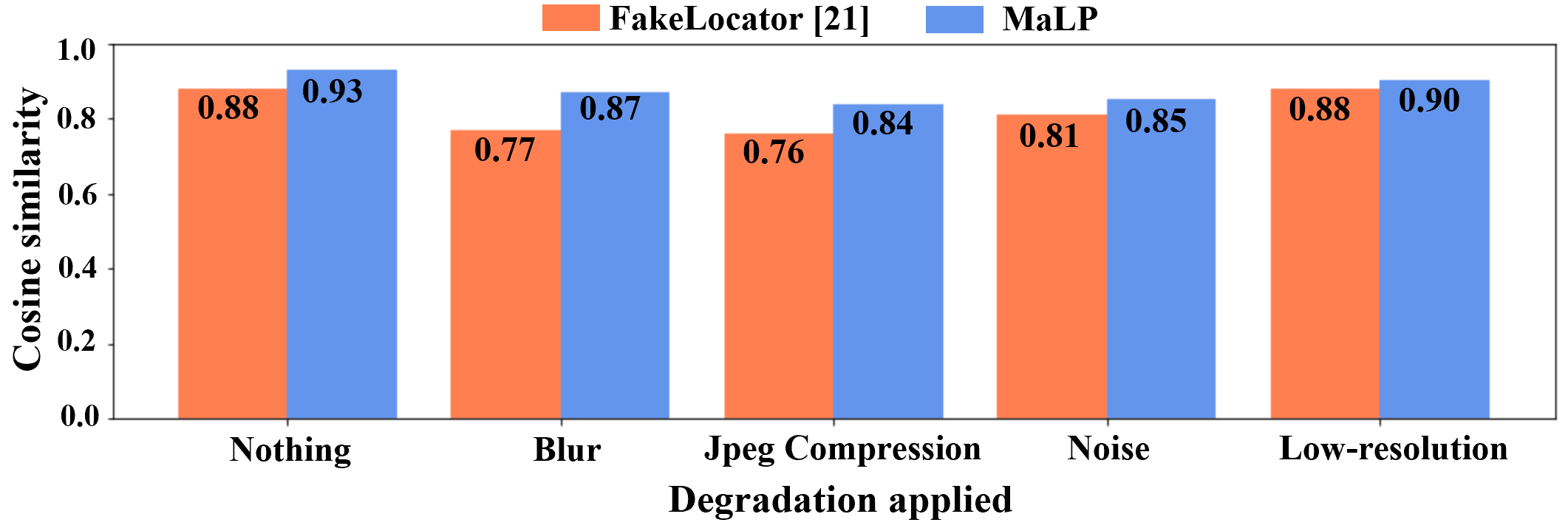}
\caption{ Comparison of our approach's robustness against common image editing degradations.  }
\label{fig:ieo}
\vspace{-3mm}
\end{figure}

\minisection{Model Robustness Against Degradations}
It is necessary to test the robustness of our proposed approach against various types of real-world image editing degradations. 
We evaluate our method on degradations applied during testing as adopted by~\cite{huang2022fakelocator}, which include JPEG compression, blurring, adding noise, and low resolution. 
The results are shown in Fig.~\ref{fig:ieo}. Our proposed MaLP is more robust to real-world degradations than passive schemes.

\subsection{Ablations}
\label{sec:ablation}
\minisection{Two-branch Architecture}
As described in Sec.~\ref{sec:localization}, MaLP adopts a two-branch architecture to predict the fakeness map using the local-level and global-level features, which are estimated by a shallow CNN and a transformer. 
We ablate by training each branch separately to show the effectiveness of combining them. As shown in Tab.~\ref{tab:architecture_ablation}, if the individual network is trained separately, the performance is lower than the two-branch architecture. 
Next, to show the efficacy of the transformer, we use a ResNet50 network in place of the transformer to predict the fakeness map. We observe that the performance is even worse than using only the transformer. ResNet50 lacks the added advantage of self-attention in the transformer, which estimates the global-level features much better than a CNN network.

\minisection{Constraints}
MaLP leverages different constraints to estimate the fakeness map using an optimized template. We perform an ablation by removing each constraint separately, showing the importance of every constraint. Tab.~\ref{tab:loss_ablation} shows the cosine similarity for localization and accuracy for detection. 
Removing either the classifier or recovery constraint results in lower detection performance. This is expected as we leverage logits from both $\mathcal{C}$ and $\mathcal{E}_E$, and removing the constraint for one network will hurt the logits of the other network. Furthermore, removing the template constraint results in a decrease in performance. 
Although the gap is small, the template is not properly optimized to have lower magnitude and high-frequency content. 

 Removing the localization constraint and just applying a $L_2$ loss for supervising fakeness maps result in a significant performance drop for both localization and detection, showing the necessity of this constraint. 
 Finally, we show the importance of a learnable template by not optimizing it during the training of MaLP. 
 This hurts the performance a lot, similar to removing the localization constraint. Both these observations prove that our localization constraint and learnable template are important components of MaLP. 

 \begin{table}[t]
\rowcolors{1}{mygray}{white}
\centering
\small
\caption{ Ablation of two-branch architecture. CNN is a shallow network with $10$ layers. Training each branch separately has worse localization results than combining them. }
\begin{NiceTabular}{c|c|c}
\hline\hline
\rowcolor{mygray}Network trained & Cosine similarity $\bf \uparrow$ & Accuracy $\bf \uparrow$\\\hline
CNN only & $0.8961$ & $0.9801$\\
\rowcolor{mygray}Transformer only & $0.8848$ & $0.9856$\\
CNN + ResNet50 & $0.8647$ & $0.9512$\\
\rowcolor{mygray}CNN $+$ Transformer & $\bf 0.9394$ & $\bf 0.9981$\\
\hline\hline
\end{NiceTabular}
\label{tab:architecture_ablation}
\vspace{-3mm}
\end{table}

\begin{table}[t]
\rowcolors{1}{mygray}{white}
\centering
\small
\caption{ Ablation of constraints used in training our framework. }
\begin{adjustbox}{width=1\columnwidth}
\begin{NiceTabular}{c|c|c}
\hline\hline
\rowcolor{mygray}Constraint removed & Cosine similarity $\bf \uparrow$ & Accuracy $\bf \uparrow$\\\hline
Classifier constraint $J_C$ & $0.9319$& $0.9814$\\
\rowcolor{mygray}Template constraint $J_T$ & $0.9143$ & $0.9803$\\
Localization constraint $J_L$ & $0.8814$& $0.9539$ \\
\rowcolor{mygray}Recovery constraint$J_R$ & $0.9206$& $0.9780$\\
Fixed template & $0.8887$ & $0.9514$\\
\rowcolor{mygray}Nothing (MaLP) & $\bf 0.9394$ & $\bf 0.9991$ \\
\hline\hline
\end{NiceTabular}
\label{tab:loss_ablation}
\end{adjustbox}
\vspace{-3mm}
\end{table}

\begin{figure}[t!]
\rowcolors{1}{mygray}{white}
\centering
\includegraphics[width=\columnwidth]{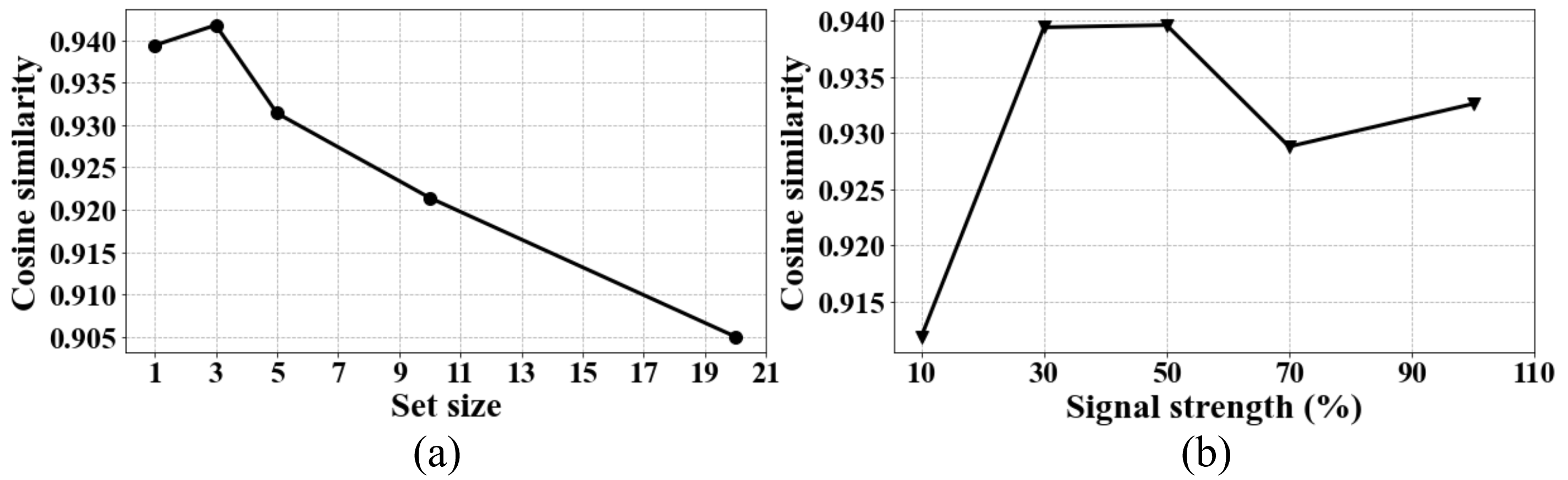}
\caption{ Ablation study on hyperparameters used in our framework: set size and signal strength.}
\label{fig:set_size_str}
\vspace{-4mm}
\end{figure}

\minisection{Template Set Size}
We perform an ablation to vary the size of the template set $\mathcal{S}$. Having multiple templates will improve security if an attacker tries to reverse engineer the template from encrypted images. The results are shown in Fig.~\ref{fig:set_size_str} (a). The cosine similarity takes a dip when the set size is increased. We also observe the inter-template cosine similarity, which remains constant at a high value of around $0.74$ for all templates. This is against the findings of~\cite{asnani2022proactive}. Localization is a more challenging task than binary detection. Therefore, it is less likely to find different templates for our MaLP in the given feature space compared to~\cite{asnani2022proactive}.  

\minisection{Signal Strength} 
We vary the template strength hyperparameter \textit{m} to find its impact on the performance. As shown in Fig.~\ref{fig:set_size_str} (b), the cosine similarity increases as we increase the strength of the added template. However, this comes with the lower visual quality of the encrypted images if the template strength is increased. The performance doesn't vary much after $m=30\%$, which we use for MaLP. 

\section{Conclusion}
This paper focuses on manipulation localization using a proactive scheme (MaLP). We propose to improve the generalization of manipulation localization across unseen GM and facial attribute modifications. We add an optimal template onto the real images and estimate the fakeness map via a two-branch architecture using local and global-level features. MaLP outperforms prior works with much stronger generalization capabilities, as demonstrated by our proposed evaluation benchmark with $22$ different GMs in various domains. We show an application of MaLP in fine-tuning GMs to improve generation quality. 

\minisection{Limitations}
First, the number of publicly available GMs is limited. More thorough testing on many different GMs might give more insights into the problem of generalizable manipulation localization. Second, we show that our MaLP can be used to fine-tune the GMs to improve image generation quality. However, it is based on the pretrained GM. Using our method to train a GM from scratch can be an interesting direction to explore in the future.

{\small
\bibliographystyle{ieee_fullname}
\bibliography{egbib}
}

\clearpage
\setcounter{equation}{0}
\setcounter{figure}{0}
\setcounter{table}{0}
\setcounter{section}{0}
\twocolumn[\centering \section*{\Large \textbf{MaLP: Manipulation Localization Using a Proactive Scheme \\ -- Supplementary material --\\[1cm]}}]

\section{Implementation Details} 

\minisection{Experimental Setup and Hyperparameters} We train MaLP for $150,000$ iterations with a batch size of $4$. 
For all of the networks, we use Adam optimizer except for the transformer which uses AdamW  with $\beta_1=0.9$, $\beta_2=0.999$, weight decay $0.5e^{-5}$ and eps $1e^{-8}$. 
The learning rate is $1e^{-5}$ for all networks. 
The constraint weights are set as: $\lambda_1=100,\lambda_2=5,\lambda_3=4,\lambda_4=25,\lambda_5=25,\lambda_6=25,\lambda_7=50,\lambda_8=15,\lambda_9=20,\lambda_{10}=50$. 
We use a template set size of $1$ and template strength as $30\%$ unless mentioned. 
All experiments are conducted on one NVIDIA K$80$ GPU. 

\begin{figure*}[t!]
\centering
\includegraphics[width=0.98\textwidth]{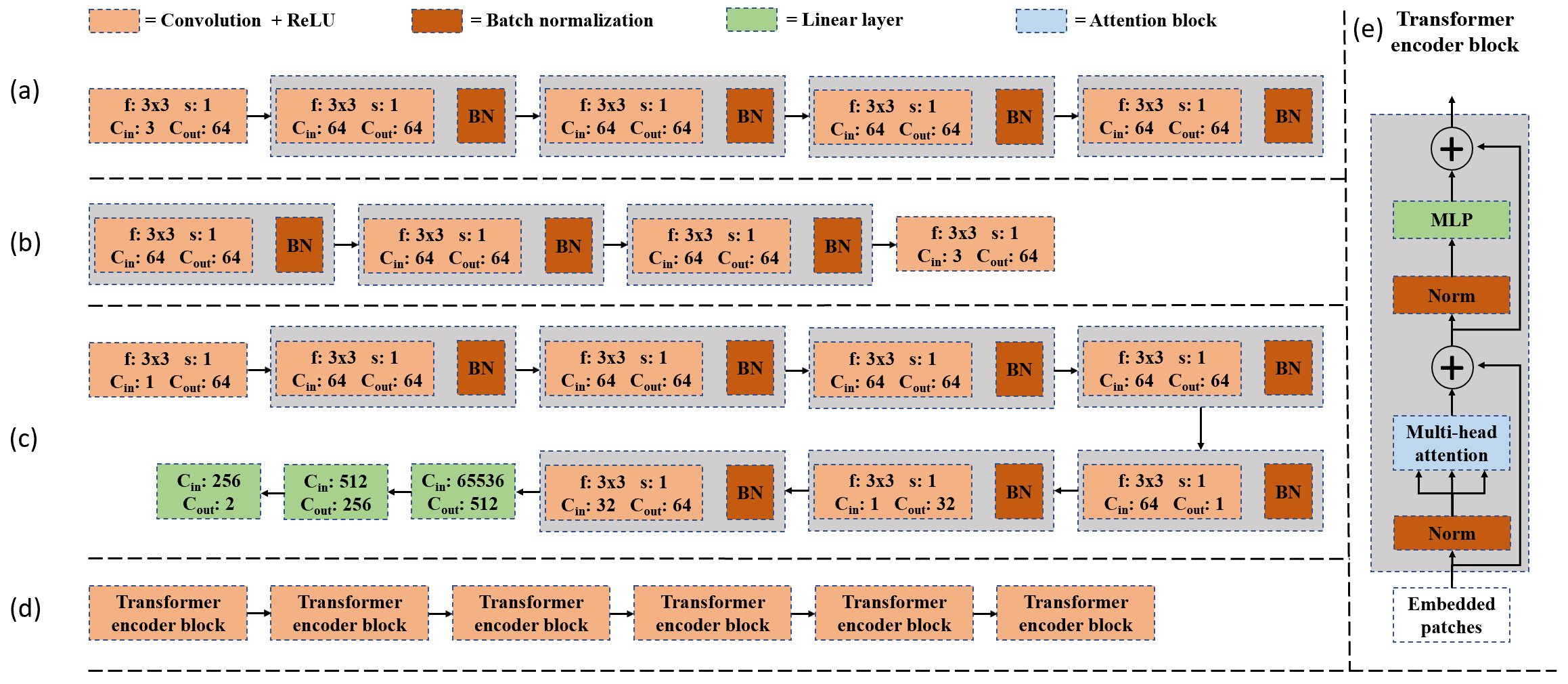}
\vspace{0.5mm}
\caption{Network architecture for different components of MaLP. (a) Shared network, (b) Encoder $\mathcal{E}_E$ and CNN network $\mathcal{E}_C$, (c) Classifier $\mathcal{C}$, (d) Transformer $\mathcal{E}_T$, and (e) Transformer encoder block.}
\label{fig:net_arc}
\vspace{-3mm}
\end{figure*}

\minisection{Network Architecture.} 
We show the network architecture of various components of MaLP in Fig.~\ref{fig:net_arc}. The shared network consists of $1$ stem convolutional layer and $4$ convolution blocks. Each convolution block consists of convolutional and batch normalization layers followed by ReLU activation. The output of the shared network is given to $\mathcal{E}_E$ and $\mathcal{E}_C$, both having the same architecture with $3$ convolution blocks and $1$ stem convolutional layer. We use the transformer $\mathcal{E}_T$ in the second branch of the framework where the ViT~\cite{dosovitskiy2020image} architecture is adopted. The transformer consists of $6$ encoder blocks, and a dropout of $0.1$ is used. The features of the transformer are reshaped to the shape of the fakeness map~\ie $1\times128\times 128$. Finally, we use a classifier $\mathcal{C}$ on the predicted fakeness maps to perform real~\vs fake binary classification. The classifier has $8$ convolution blocks, $1$ stem convolutional layer, and $3$ fully connected layers. We apply the ReLU activation between the layers. 

\begin{figure*}[t!]
\centering
\includegraphics[width=0.98\textwidth]{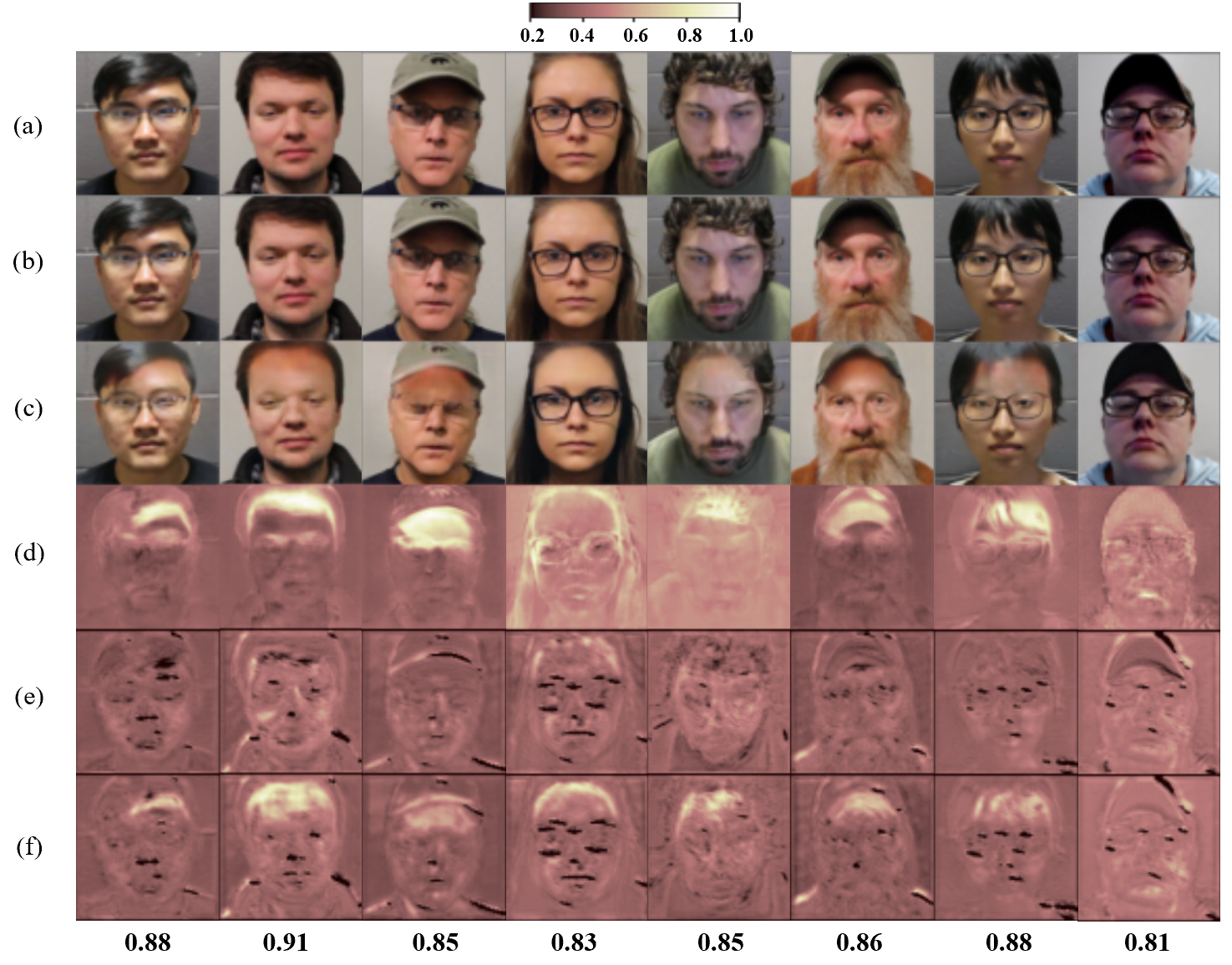}
\vspace{0.5mm}
\caption{Visualization of fakeness maps for different attribute modifications by STGAN. (a) Real image, (b) encrypted image, (c) manipulated image, (d) ground-truth $\vect{M}_{GT}$, (e) predicted fakeness map for encrypted images, and (f) predicted fakeness map for manipulated images. We also show the cosine similarity between the predicted and ground-truth fakeness map below (f). All face images come from SiWM-v2 data~\cite{swim}.}
\label{fig:vis_supp_1}
\vspace{-3mm}
\end{figure*}

\begin{figure*}[t!]
\centering
\includegraphics[width=0.98\textwidth]{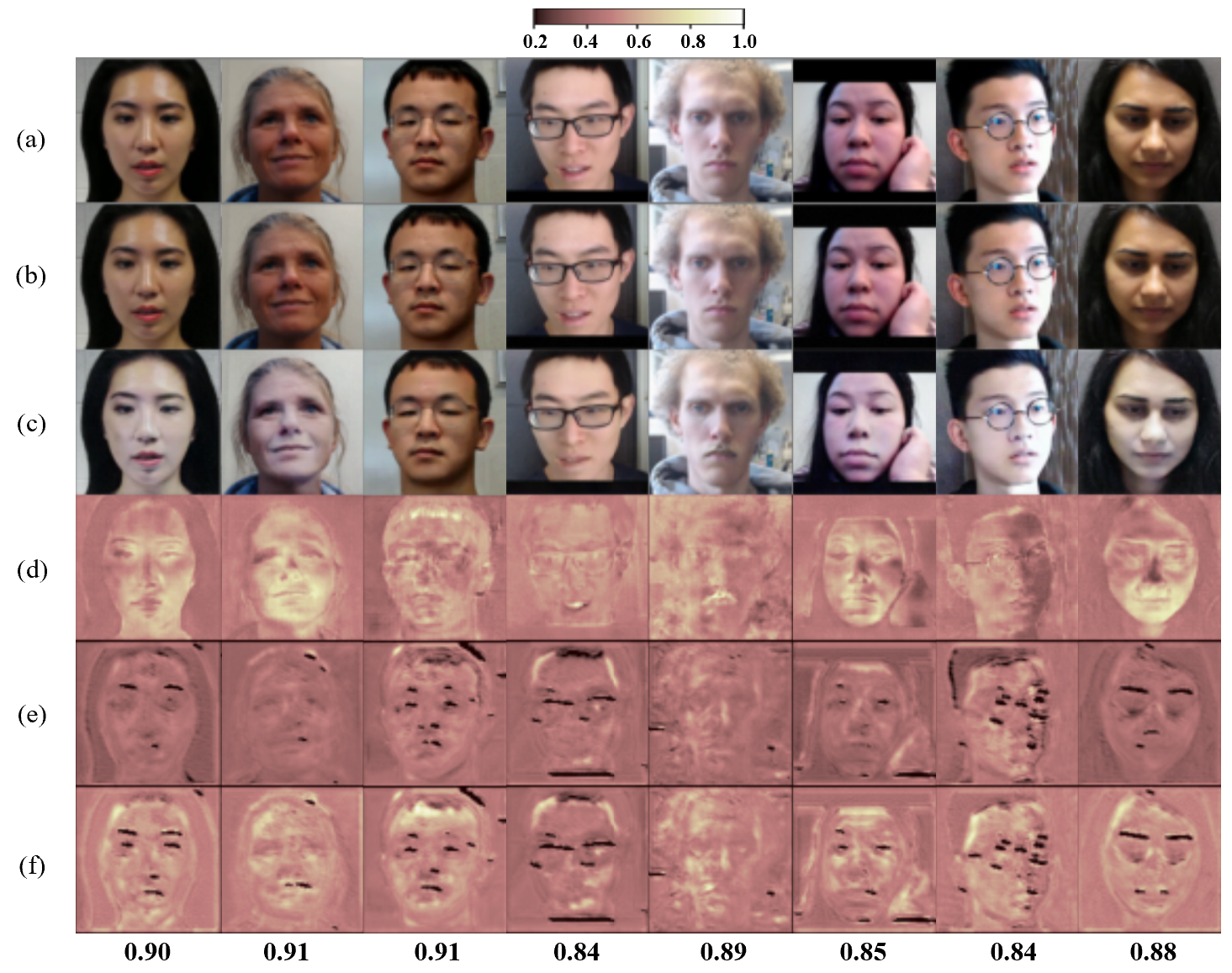}
\vspace{0.5mm}
\caption{Visualization of fakeness maps for different attribute modifications by STGAN. (a) Real image, (b) encrypted image, (c) manipulated image, (d) ground-truth $\vect{M}_{GT}$, (e) predicted fakeness map for encrypted images, and (f) predicted fakeness map for manipulated images. We also show the cosine similarity between the predicted and ground-truth fakeness map below (f). All face images come from SiWM-v2 data~\cite{swim}.}
\label{fig:vis_supp_2}
\vspace{-3mm}
\end{figure*}

\begin{figure*}[t!]
\centering
\includegraphics[width=0.98\textwidth]{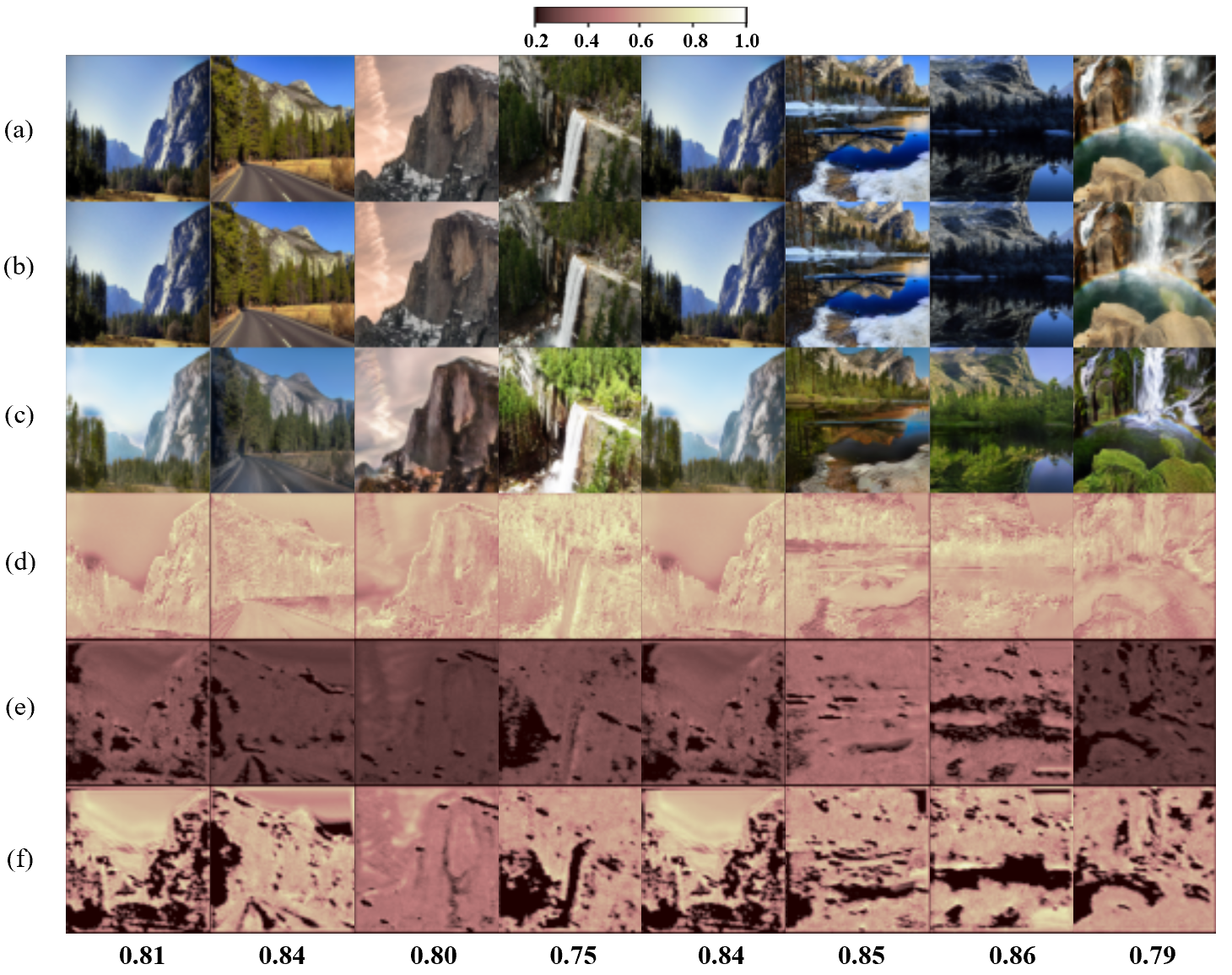}
\vspace{0.5mm}
\caption{Visualization of fakeness maps for manipulation by DRIT. (a) Real image, (b) encrypted image, (c) manipulated image, (d) ground-truth $\vect{M}_{GT}$, (e) predicted fakeness map for encrypted images, and (f) predicted fakeness map for manipulated images. We also show the cosine similarity between the predicted and ground-truth fakeness map below (f).}
\label{fig:vis_supp_3}
\vspace{-3mm}
\end{figure*}

\begin{figure*}[t!]
\centering
\includegraphics[width=0.98\textwidth]{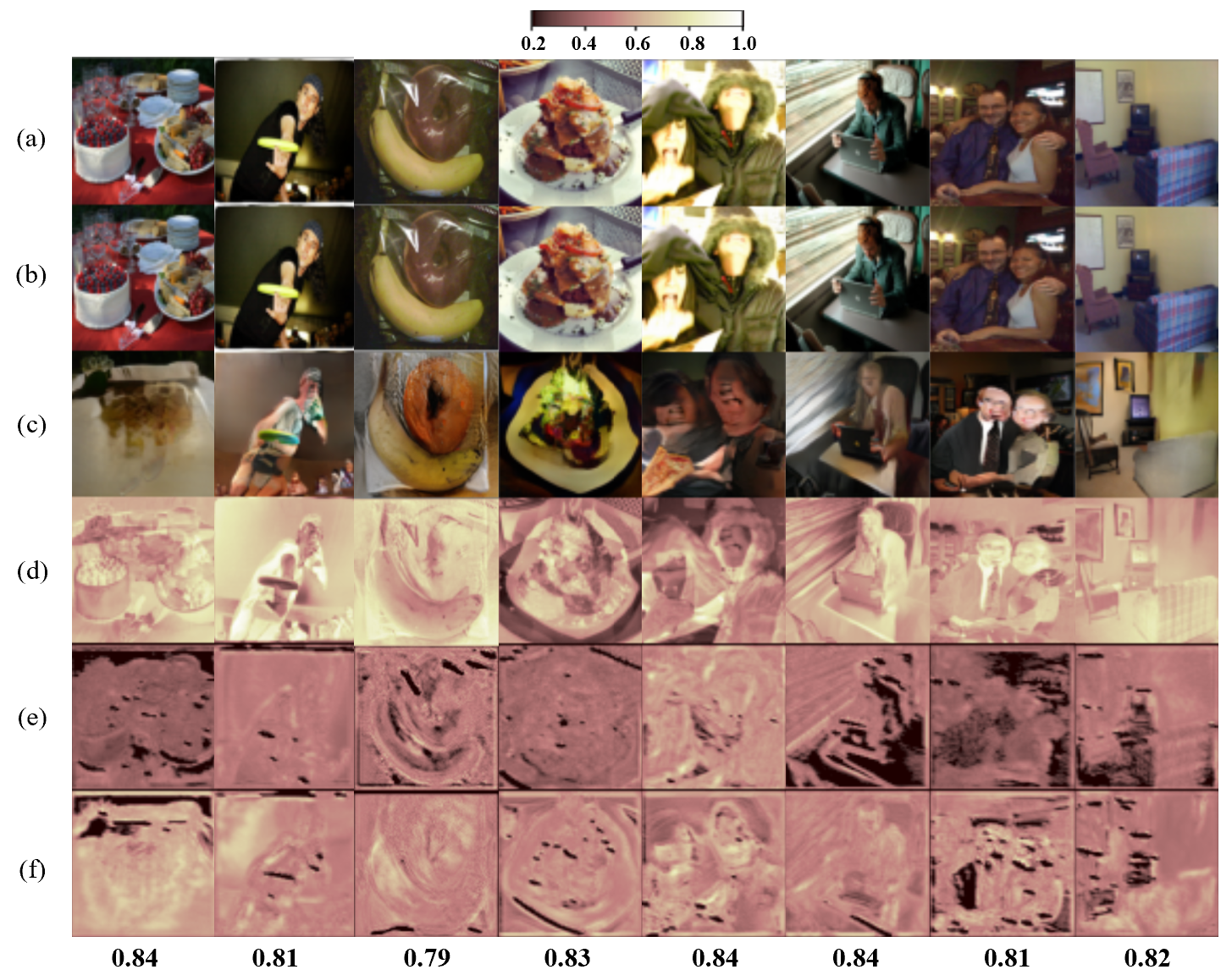}
\vspace{0.5mm}
\caption{Visualization of fakeness maps for manipulation by GauGAN. (a) Real image, (b) encrypted image, (c) manipulated image, (d) ground-truth $\vect{M}_{GT}$, (e) predicted fakeness map for encrypted images, and (f) predicted fakeness map for manipulated images. We also show the cosine similarity between the predicted and ground-truth fakeness map below (f).}
\label{fig:vis_supp_4}
\vspace{-3mm}
\end{figure*}

\minisection{GMs and dataset license information.} 
We use a variety of face and generic GMs to show the effectiveness of MaLP. The information for all the GMs along with their training datasets, is shown in Tab.~\ref{tab:dataset_gms}. For many GMs used by~\cite{asnani2022proactive}, We use the test images released by~\cite{asnani2022proactive}. for the remaining GMs, we would release the test images for fair comparison of generalization benchmark by the future works. We also show more visualization samples of the predicted fakeness maps by MaLP in Fig.~\ref{fig:vis_supp_1}-~\ref{fig:vis_supp_4}. All the fakeness maps are shown in "pink" cmap for better representation. We also indicate the cosine similarity between the predicted and ground truth fakeness maps. We observe that the fakeness maps for encrypted images have minimal bright regions. However, for fake images, MaLP is able to localize the modified regions well, considering the modified attributes/GMs are unseen in training. 

The face datasets include CelebA~\cite{liu2018large} and CelebA-HQ~\cite{karras2018progressive}, both of which don't have any associated Institutional Review Board (IRB) approval. The authors for both datasets mention the availability of the dataset for non-commercial research purposes, which we strictly adhere to. For generic images datasets, we use Facades~\cite{Tylecek13}, COCO~\cite{coco}, Horse2Zebra~\cite{CycleGAN2017}, Summer2Winter~\cite{CycleGAN2017}, GTA2CITY~\cite{Richter_2016_ECCV}, Edges2Shoes~\cite{isola2017pix2pix}, Paris street-view~\cite{pathakCVPR16context} and Sketch-Photo~\cite{wang2008face} datasets. All the mentioned generic image datasets can be used for non-commercial research purposes, as mentioned by the authors, and we use the datasets for the same purposes.

\begin{table}[t]
\rowcolors{1}{mygray}{white}
\begin{center}
\small
\centering{
\caption{List of GMs along with their training datasets}
\begin{adjustbox}{width=1\columnwidth}
\label{tab:dataset_gms}
\begin{NiceTabular}{c|c}
\hline\hline
\rowcolor{mygray} Dataset & GMs \\\hline
& STGAN~\cite{liu2019stgan}, AttGAN~\cite{he2019attgan}, StarGAN~\cite{choi2018stargan},\\
& GANimation~\cite{pumarola2018ganimation}, CouncilGAN~\cite{nizan2020council},\\
\multirow{-3}{*}{CelebA~\cite{liu2018large}} &   ESRGAN~\cite{wang2021realesrgan}, GDWCT~\cite{GDWCT2019}\\
\rowcolor{mygray}& SEAN~\cite{Zhu_2020_sean}, StarGAN-v2~\cite{stargan2}, ALAE~\cite{pidhorskyi2020alae},\\
\rowcolor{mygray}\multirow{-2}{*}{CelebA-HQ~\cite{karras2018progressive}} &  DRGAN~\cite{tran2017disentangled}, ColorGAN~\cite{nazeri2018image},  \\
Facades~\cite{Tylecek13} & CycleGAN~\cite{CycleGAN2017}, BicycleGAN~\cite{zhu2017bicycle}, Pix2Pix~\cite{isola2017pix2pix}\\
\rowcolor{mygray}COCO~\cite{coco} & GauGAN~\cite{park2019gaugan} \\
Horse2Zebra~\cite{CycleGAN2017} & AutoGAN~\cite{zhang2019detecting} \\
\rowcolor{mygray}Summer2Winter~\cite{CycleGAN2017} & DRIT~\cite{DRIT} \\
GTA2CITY~\cite{Richter_2016_ECCV} & UNIT~\cite{liu2018unit}\\
\rowcolor{mygray}Edges2Shoes~\cite{isola2017pix2pix} & MUNIT~\cite{huang2018munit}\\
Paris Street-view~\cite{nizan2020council} & Cont\_Enc~\cite{pathakCVPR16context}\\
\rowcolor{mygray}Sketch-Photo~\cite{wang2008face} & DualGAN~\cite{yi2018dualgan}\\
\hline\hline
\end{NiceTabular}
\end{adjustbox}}
\end{center}
\vspace{-3mm}
\end{table}

\minisection{Image Editing Degradations.}
We apply several image editing degradations to the test set to verify the robustness of MaLP. The details of these operations are listed below:
\begin{enumerate}
    \item JPEG compression: We compress the image with the compression quality of $50\%$.
    \item Blur: We apply the Gaussian blur with a filter size of $7\times7$.
    \item Noise: We apply a Gaussian noise with zero mean and unit variance.
    \item Low-resolution: We resize the image to half the original resolution and restore it back to the original resolution using linear interpolation. 
    
\end{enumerate}

\minisection{Potential Societal Impact} The problem of manipulation localization is crucial from the perspective of media forensics. Localizing the fake regions not only helps in the detection of these fake media but, in the future, can also help recover the original image that the GM has manipulated. We also show that MaLP can be used as a discriminator to improve the quality of GMs. While this is an interesting application of MaLP, it can be a possibility that the GMs become more robust to our framework, decreasing the localization performance if the training of the GM is done from scratch.

\begin{table}[t]
\rowcolors{1}{mygray}{white}
\begin{center}
\centering{
\caption{Ablation for localization loss.}
\begin{adjustbox}{width=0.8\columnwidth}
\tiny
\label{tab:man_loss_ablation}
\begin{NiceTabular}{c|c|c|c}
\hline\hline
\rowcolor{mygray} Loss & CS $\bf \uparrow$ & PSNR $\bf \uparrow$ & SSIM $\bf \uparrow$\\\hline
 CS & $0.9356$ & $22.16$ & $0.7114$\\
\rowcolor{mygray} CS $+$ $\text{L}_2$ & $0.9230$ & $18.98$ & $0.6614$\\
 CS $+$ SSIM $+$ $\text{L}_2$& $0.9211$ & $19.12$ & $0.6816$\\
\rowcolor{mygray} CS $+$ SSIM $+$ $\text{L}_1$ & $0..8777$ & $14.01$ & $0.3712$\\
 CS $+$ SSIM  & $\bf 0.9394$ & $\bf 23.020$ & $\bf 0.7312$ \\\hline\hline
\end{NiceTabular}
\end{adjustbox}}
\end{center}
\vspace{-3mm}
\end{table}

\section{Additional Experiments}

\minisection{Localization Loss.} We show the importance of manipulation loss (defined in Eq. $8$) in Sec. $4.6$. We perform an ablation to formulate the loss of fakeness maps for manipulated images. As shown in Tab.~\ref{tab:man_loss_ablation}, we try experimenting with various loss functions~\ie cosine similarity (CS), $\text{L}_1$, $\text{L}_2$ and structural similarity index measure (SSIM). Using just the CS loss results in better performance compared to combining it with $\text{L}_1$ or $\text{L}_2$ loss. We observe a huge deterioration in performance when using $\text{L}_1$ loss. This can be explained as PSNR and SSIM are directly related to mean squared error which is optimized by either an $\text{L}_2$ or SSIM loss. Finally, adopting an SSIM loss with CS loss results in a better performance as both of them are more related to the metrics, making it easier for MaLP to converge.

\begin{table}[t]
\rowcolors{1}{mygray}{white}
\begin{center}
\small
\centering{
\caption{Comparison with~\cite{huang2022fakelocator} using multiple GMs in training. MaLP is able to outperform~\cite{huang2022fakelocator} by training images manipulated by only STGAN.}
\begin{adjustbox}{width=1\columnwidth}
\label{tab:baseline_comp_multi_gm}
\begin{NiceTabular}{c|c|c|c|c}
\hline\hline
\rowcolor{mygray}& & \multicolumn{3}{c}{Cosine similarity $\bf \uparrow$}\\\hhline{~|~|-|-|-}
\rowcolor{mygray} \multirow{-2}{*}{Method} & \multirow{-2}{*}{Training GMs} & AttGAN & StarGAN & StyleGAN\\\hline
 & STGAN $+$ ICGAN $+$ PGGAN & & & \\
& $+$ StyleGAN $+$ StyleGAN2 & & & \\
\multirow{-3}{*}{Hunag~\etal~\cite{huang2022fakelocator}} & $+$ StarGAN $+$ AttGAN & \multirow{-3}{*}{$0.6940$}& \multirow{-3}{*}{$0.8494$}& \multirow{-3}{*}{$0.7479$}\\
\rowcolor{mygray}MaLP & STGAN & $\bf 0.8557$ & $\bf 0.8718$ & $\bf 0.8255$\\\hline\hline
\end{NiceTabular}
\end{adjustbox}}
\end{center}
\vspace{-3mm}
\end{table}

\begin{table}[t]
\rowcolors{1}{mygray}{white}
\begin{center}
\small
\centering{
\caption{Performance of MaLP across different attribute modifications seen in training.}
\begin{adjustbox}{width=1\columnwidth}
\label{tab:baseline_comp_multi_att}
\begin{NiceTabular}{c|c|c|c|c|c|c}
\hline\hline
\rowcolor{mygray}& \multicolumn{6}{c}{Cosine similarity $\bf \uparrow$}\\\hhline{~|-|-|-|-|-|-}
\rowcolor{mygray} \multirow{-2}{*}{Method} & Bald & Bangs & Black Hair & Eyeglasses & Mustache & Smile\\\hline
\cite{huang2022fakelocator} & $0.9014$ & $0.8850$ & $0.8817$ & $0.9093$ & $0.9152$ & $0.8634$\\
\rowcolor{mygray}MaLP & $\bf 0.9478$& $\bf 0.9329$& $\bf 0.9367$& $\bf 0.9549$& $\bf 0.9470$& $\bf 0.9489$\\\hline\hline
\end{NiceTabular}
\end{adjustbox}}
\end{center}
\vspace{-3mm}
\end{table}

\minisection{Comparison with Baseline.} 
Due to the limited GPU memory, we conduct proactive training with one GM only because the GM needs to be loaded to the memory and used on the fly. On the other hand, passive methods can be trained on multiple GMs because the image generation processes are conducted offline.
As shown in Tab.~\ref{tab:baseline_comp_multi_gm},~\cite{huang2022fakelocator} trains on images manipulated by $7$ different GMs, unlike MaLP, which is trained on images manipulated by only $1$ GM. We show the performance on three GMs, which are seen for~\cite{huang2022fakelocator}, but unseen for MaLP. MaLP performs better even though these GMs' images are not seen in training. Therefore, even though the training of MaLP is limited by $1$ GM, it can achieve better generalization to other GMs proving the effectiveness of proactive schemes.

\minisection{Multiple Attribute Modifications.} 
Instead of training on bald attribute modification by STGAN, we train and test MaLP on multiple attribute modifications. These include bald, bangs, black hair, eyeglasses, mustache, and smile manipulation. We show the results in Tab.~\ref{tab:baseline_comp_multi_att}. MaLP performs better for all the attribute modifications compared to the passive method~\cite{huang2022fakelocator}. We also observe an increase in cosine similarity compared to when MaLP is trained on only bald attribute modification. This is expected, as the more types of modifications MaLP sees in training, the better it learns to localize. 

\begin{table}[t]
\rowcolors{1}{mygray}{white}
\begin{center}
\small
\centering{
\caption{Ablation study for transformer architecture.}
\begin{adjustbox}{width=1\columnwidth}
\label{tab:trans_arc_abl}
\begin{NiceTabular}{c|c|c|c|c}
\hline\hline
\rowcolor{mygray} Optimizer & Depth & Dropout & Cosine similarity$\bf \uparrow$ & Accuracy$\bf \uparrow$ \\\hline
Adam & $6$ & $0.1$  & $0.8839$ & $0.9514$\\
\rowcolor{mygray}AdamW & $1$ & $0.0$  & $0.8825$ & $0.9647$\\
AdamW & $1$ & $0.0$  & $0.8826$ & $0.9680$\\
\rowcolor{mygray}AdamW & $3$ & $0.0$ & $0.8830$ & $0.9705$\\
AdamW & $6$  & $0.1$ & $\bf0.8848$ & $\bf0.9856$\\
\hline\hline
\end{NiceTabular}
\end{adjustbox}}
\end{center}
\vspace{-3mm}
\end{table}

\minisection{Transformer Architecture Ablation.} We ablate various parameters of the transformer to select the best architecture for manipulation localization. We experiment with parameters that include optimizer, depth~\ie number of blocks, and dropout. We only use the transformer branch and switch off the CNN branch during training. The results are shown in Tab.~\ref{tab:trans_arc_abl}. We observe that the localization performance is almost the same when using the transformer to predict fakeness maps. However, the detection accuracy has a significant impact. Having dropout does increase the performance for detection and localization. Further, using the weighted Adam optimizer is more beneficial than using the vanilla Adam optimizer. Therefore, we adopt the architecture of the transformer with $6$ blocks and optimize it with a weighted Adam optimizer. Finally, we also include the dropout to achieve the best performance for localization and detection.

\end{document}